%% file: ms.tex
\documentclass[a4paper,fleqn]{cas-dc}

\usepackage{graphicx}
\usepackage{graphics}
\usepackage{amssymb}
\usepackage{hyperref}
\usepackage{graphicx}
\usepackage{color}
\usepackage{booktabs}
\usepackage{bigstrut}
\usepackage{multirow}
\usepackage{array}
\usepackage{tabularx}
\usepackage{algorithm}
\usepackage[]{mdframed}
\usepackage{algpseudocode}
\usepackage{amsmath}
\usepackage{amssymb}
\usepackage{pifont}
\usepackage{tikz}
\usepackage{subfig}

\usepackage{tikz}
\usepackage{array}
\usepackage{amsmath}
\usetikzlibrary{arrows,positioning,shapes.geometric}
\usepackage[utf8]{inputenc} 
\usepackage[T1]{fontenc}    
\usepackage{hyperref}       
\usepackage{url}            
\usepackage{booktabs}       
\usepackage{amsfonts}       
\usepackage{nicefrac}       
\usepackage{microtype}      
\usepackage{lipsum}
\usepackage{fancyhdr}       
\usepackage{graphicx}       
\usepackage{longtable}
\usepackage{caption} 
\graphicspath{{media/}}     
\usepackage{xcolor,colortbl}
\definecolor{Gray}{gray}{0.82}
\definecolor{Gray1}{gray}{0.89}
\usepackage{lipsum}

\usepackage[square,sort,comma,numbers]{natbib}

\def\tsc#1{\csdef{#1}{\textsc{\lowercase{#1}}\xspace}}
\tsc{WGM}
\tsc{QE}

\begin{document}
\let\WriteBookmarks\relax
\def\floatpagepagefraction{1}
\def\textpagefraction{.001}


\shorttitle{Online Stock Forecasting Dataset}    

\shortauthors{Bathini, Cihan}  

\title [mode = title]{Real-Time Online Stock Forecasting Utilizing Integrated Quantitative and Qualitative Analysis}  


%

\author{Bathini Sai Akash}[orcid=0000-0001-7408-6925]

\cormark[1]


\ead{f20190065h@alumni.bits-pilani.ac.in}

\ead[url]{https://orcid.org/0000-0001-7408-6925}

\credit{Conceptualization, Writing - Original Draft, Software, Methodology, Formal analysis, Validation, Investigation}

\affiliation{organization={BITS Pilani Hyderabad Campus},
            addressline={Secunderabad}, 
            city={Hyderabad},
            postcode={500084}, 
            state={Telangana},
            country={India}}

\author{Cihan Dagli}[orcid=0000-0003-4919-1699]


\ead{dagli@mst.edu}

\ead[url]{https://works.bepress.com/cihan-dagli/}

\credit{Funding acquisition, Supervision, Conceptualization, Resources, Project administration }

\affiliation{organization={Missouri University of Science and Technology},
            addressline={Parker Hall, 106, 300 W 13th St}, 
            city={Rolla},
            postcode={65409}, 
            state={Missouri},
            country={USA}}
\cortext[1]{Corresponding author}



\begin{abstract}
The application of Machine learning to finance has become a familiar approach, even more so in stock market forecasting. The stock market is highly volatile, and huge amounts of data are generated every minute globally. The extraction of effective intelligence from this data is of critical importance. However, a collaboration of numerical stock data with qualitative text data can be a challenging task. In this work, we accomplish this by providing an unprecedented, publicly available dataset with technical and fundamental data and sentiment that we gathered from news archives, TV news captions, radio transcripts, tweets, daily financial newspapers, etc. The text data entries used for sentiment extraction total more than 1.4 Million. The dataset consists of daily entries from January 2018 to December 2022 for eight companies representing diverse industrial sectors and the Dow Jones Industrial Average (DJIA) as a whole. Holistic Fundamental and Technical data is provided training ready for Model learning and deployment. Most importantly, the data generated could be used for incremental online learning with real-time data points retrieved daily since no stagnant data was utilized. All the data was retired from APIs or self-designed robust information retrieval technologies with extremely low latency and zero monetary cost. These adaptable technologies facilitate data extraction for any stock. Moreover, the utilization of  Spearman's rank correlation over real-time data, linking stock returns with sentiment analysis has produced noteworthy results for the DJIA and the 8 other stocks, achieving accuracy levels surpassing 60\%. The dataset is made available at \href{https://github.com/batking24/Huge-Stock-Dataset}{Github}.
\end{abstract}


\begin{highlights}
\item A robust dataset, with quantitative and qualitative data for Incremental Online learning at stock forecasting.
\item Incorporated comprehensive technical and fundamental analysis literary review. 
\item BERT-based language models for sentiment analysis, extensive comparative analysis
\item Spearman correlation of stock return with extracted sentiment
\item Relevant information retrieval tools for stock-specific data extraction
\end{highlights}


\maketitle





\input{Sections/Introduction.tex}

\input{Sections/related_work_dataset.tex}

\input{Sections/Dataset_Description.tex}

\input{Sections/Appendix}

\printcredits




\bibliographystyle{unsrt}  
\bibliography{references}

\end{document}

%% file: Sections/Introduction.tex
\section{Introduction}
Stock market forecasting is an ever-green field with extensive research being performed consistently. Early research was established on EMH (Efficient Market Hypothesis) which states that stock share prices reflect and capture all information. Further research in the field claimed otherwise\cite{ZHANG2018236}. Stock prices are highly volatile because of numerous diversified factors and events. These include investor sentiment and behavior, economic factors, leadership, and political factors, policy changes, natural disasters, etc \cite{zouaoui2011does}\cite{otoo1999consumer}. However, this has proven to be insufficient.

By and Large, data for stock forecasting can be differentiated into two categories, quantitative and qualitative data. Quantitative data is historical stock-related data and macroeconomic features. Qualitative data accounts for investors' reactions to daily events in the stock market, company management and strategies, balance sheets, company leadership, natural disasters, etc.

Long-established stock market forecasting techniques made use of historical stock data which is available through numerous sources like Yahoo Finance, Bloomberg, Wind, etc but in the Big data era with the significant progress of both information communication and Natural language processing, user-originated traditional data such as discussion forums, tweets, news articles, earning reports, and analytic reports has become paramount. \cite{sun2016trade}.

Studies have shown that there is a strong correlation between stock forecasting and online news articles\cite {schumaker2009textual}. Financial reviews and expert discussions influence people's behavior in financial markets. Therefore, it's vital that both qualitative and quantitative data be effectively combined.\cite{agarwal2019stock}
Given this, studies show that only 11\% out of 122 studies in stock forecasting proposed a collaboration of qualitative and quantitative data. Since the stock market is an umbrella market, dependence on one specific data source might not yield accurate predictions.

However, there are challenges to extracting information from sparse web data. First, getting benchmark, recognized data is difficult. Even though the availability of text data is soaring, it is usually limited to prominent stocks and might not be of a good standard. Additionally, data is sparse meaning getting daily sentiments for stocks can prove to be difficult. Further, extraction has to be performed from unstructured web sources which are restricted by IP blocking. Next, both events and emotions contribute to a qualitative understanding of financial markets. Blooming progress in NLP has given researchers a new purview on stock market prediction by extraction of investor emotions and moods\cite{chun2021using}\cite{7563727}. 

It is arduous and impractical for investors or researchers to analyze big data for predictions. Added to this there is a heavy requirement of domain knowledge in finance for the selection of data parameters. The purpose of the paper is to provide a publicly available data set for stock market prediction using qualitative data that is ready for ML model learning that could be implemented by the layman Machine Learning research community. The dataset comprises 8 stocks from diverse industries. Added to this, Dow Jones Industrial Average(DJIA) data is also provided. In our dataset, we make use of financial articles from recognized sources online to mitigate fake news.
The main contributions of the study to the research community are summarised as follows:

\begin{itemize}
    \item[1] An extensive dataset with Quantitative and Qualitative Data which is training ready for comprehensive research and studies.
    \item[2] Exhaustive Technical and Fundamental analysis review from a broad range of papers and inclusion of these indicators in our dataset
    \item[3] Substantial data collection sources and techniques including APIs, twitter data, radio transcripts, and news archive data. 
    \item[4] Implementation of multiple BERT-based Language models pre-trained on the collected dataset and Fine-tuned for sentiment analysis task. Comparative analysis of all models used. 
    \item[5] Spearman correlation analysis of the sentiment extracted through sentiment extraction pipeline utilizing proposed Language models, with stock returns

\end{itemize}

%% file: Sections/related_work_dataset.tex
\section{Related Work}
\vspace{8pt}
Early proceedings debated the possibility of predicting the stock market. According to the Random Walk theory, stock prices move stochastically in a random walk, making it impossible to forecast and indicating that any attempt at prediction would be unsuccessful. \cite{malkiel1973random}. Next, the efficient market hypothesis states that the market is random but also efficient in determining the optimal stock market price. However, further research in the field has shown that stock price direction analysis is possible.\cite{de1995financial}\cite{jegadeesh1993returns}
However, as raised before historical stock data has limitations in unfolding all the 
information relating to firms' financial well-being.

Research was conducted by making use of numerous text sources and indexes. \cite{mao2012correlating}\cite{mao2013twitter}\cite{schumaker2009textual} extracted stock text news, twitter data for S\&P500 index. The accuracy reported for stock movement prediction in \cite{mao2012correlating} was 68\%.  \cite{vu2012experiment} extracted data for Nasdaq stocks with a mean accuracy of 77\% delivered over four stocks. In \cite{bollen2011twitter} tweets were collected for Dow Jones Industrial Average(DJIA). The work further used a Deep Neural network to predict DJIA daily price movement with 88\% accuracy. Some research was performed on specific companies too \cite{kaya2010stock}. Similarly in \cite{roy2015stock} thirteen-year Goldman Sachs Group stock data was collected. In \cite{nassirtoussi2015text} Forex currency price and related news data were utilized. Finally, in \cite{patel2015predicting} data on stock prices were gathered from the CNX Nifty and S\&P BSE Sensex exchanges. 

 In this study, forecasting and predictions are used interchangeably. Stock market prediction was also based on different time frames. These time frames can be established for intraday or inter-day trading. Most research aimed at making forecasting for stock movement for one next day \cite{mao2012correlating}\cite{vu2012experiment}\cite{bollen2011twitter}. Gong et al. in \cite{gong2009new} aimed at making monthly directional predictions. Patel et al. in \cite{patel2015predicting} made predictions on a range of time frames from 1 to 30 days. Nassirtoussi et al. in \cite{nassirtoussi2015text} implemented predictions every 20 minutes.

As mentioned, the analysis of stocks happens through two categories, qualitative and quantitative. Quantitative analysis includes two sub-categories: Technical indicators and Quantitative Fundamental Analysis. Quantitative analysis typically consists of Qualitative Fundamental analysis. The following section elucidates some current and pertinent literature based on each category. 

 
\subsection{Quantitative  Based Studies}
\vspace{7pt}
\subsubsection{Technical Indicators }
\vspace{5pt}
Technical analysis of stocks aims to understand stock behavior by analyzing statistical trends, patterns, and historical market data.  It is founded on the idea that a financial time series' future behavior is conditioned by its own history.
In the wide range of research performed on Qualitative data, different studies make use of different technical indicators\cite{leigh2002forecasting}, partly because of different indexes used. 
Some studies used common technical indicators in stock markets \cite{kirkpatrick2010technical}, while others used a more diverse range of indicators\cite{zhai2007combining}. In this work, we perform an extensive review of which technical indicators are more effective in relation to different indexes and comprehend the best indicators to be included in our Dataset. The review is demonstrated in Table(\ref{table2}). 

Further, Table(\ref{table1}) defines different stock Technical Indicators along with the symbols used in our Literary review.
\\

\subsubsection{Quantitative Fundamental analysis}
\vspace{5pt}
Fundamental analysis of stocks is a method used by investors to evaluate the intrinsic value of a company by examining its underlying financial and qualitative factors. Quantitative Fundamental analysis is based on the usage of numerical data, and Qualitative Fundamental analysis makes use of text data.
Technical analysis tactics simply take into account a stock's price history; Quantitative fundamental analysis techniques take into account a company's financial condition as well as macroeconomic variables \cite{abarbanell1998abnormal}. Even though advocates of EMH contend that a stock's intrinsic value is always equal to its present price, fundamental analysts purchase and sell companies when the intrinsic value is lower or higher than the market price.

One of the underlying assumptions of fundamental analysis is that, in the case of publicly available financial data, the stock's current price often misrepresents the company's value. A second hypothesis is that the value that is obtained from the fundamental data of the firm will most likely be closer to the stock value. The financial information made available by the company whose stock is being studied forms the basis of fundamental research. Ratios and other indicators that demonstrate a company's performance relative to other enterprises of a similar size are computed using the data. To determine if the markets have overvalued or undervalued a company or asset, stock analysts use fundamental research as an evaluation tool.

To estimate and evaluate future stock prices, Quantitative fundamental analysis considers the financial health of the company, financial statements, its Board of Directors, its workers, annual reports, geographic and weather patterns- factors like natural catastrophes, as well as cash flows and balances, and income statements and geopolitical/governmental data. 

Studies have concluded that regional macroeconomic factors also have a significant impact on how the equity market performs. According to Zhao et al. \cite{zhao2010dynamic}, the trajectory of the A-share markets in mainland China can be impacted by changes in the RMB exchange rate. Consequently, including macroeconomic variables will provide more intelligence for neural network training for better training.

However, it turns out that macroeconomic factors specifically have not received much recognition in being used for stock market predictions as shown in Table(\ref{table4}) despite the fact that there is a positive correlation between macroeconomic variables and stock-market returns\cite{nti2020systematic}.

There are two methods for fundamental analysis: top-down and bottom-up. 
The \textit{top down} method is an investing strategy that begins with an examination of the economy as a whole, followed by an examination of the many industries and businesses that make up each one.
In \textit{Bottom up} approach, Investors initially concentrate on a specific firm, investigating the business model and development possibilities.

Typically, fundamental research is conducted from “a macro to the micro viewpoint” to pinpoint securities/assets that the marketplace has not accurately valued. 
Analysts often research the following topics in the following order: 
\begin{itemize}
\item The situation of the broader economy
\item The health of a particular industry
\item  The financial results of the firm issuing the stock.

\end{itemize}
If done effectively, ideally this guarantees they determine the stock's fair market value.

Automation of fundamental analysis is challenging because fundamental elements are ad hoc in nature. However, scholars/researchers can now automate stock market prediction using unstructured data thanks to the development of deep learning, which in certain circumstances has led to greater predictive performance. Fundamental analysis is beneficial for predicting long-term changes in stock prices, but it is not appropriate for predicting sudden changes in stock prices. 

In our study, we made use of Macroeconomic factors, (shown under \colorbox{green!20}{Green} in Table(\ref{table3})) and Financial Ratios. Financial Ratios of stocks refer to the quantitative metrics or factors associated with a company's financial performance and position. All the different ratios and Macroeconomic factors are explained in Table(\ref{table3}) with color coding for readability. 
Different Financial Ratios can be categorized into(\cite{baresa2013strategy}): 
\begin{itemize}
    \item[1] \textit{Liquidity Ratios}: an estimate of a company's capacity to fulfil its upcoming short-term obligations. Indicators of liquidity reveal a company's capacity to avoid bankruptcy. \colorbox{pink!30}{Pink}
    \item[2] \textit{Financial Leverage Ratios}: quantify the amount derived from foreign origins. Displays the ratio of assets funded by equity from shareholders to assets financed by liabilities.\colorbox{violet!30}{Violet}
    \item[3] \textit{Profitability Ratios}: Profitability ratios display the firm's financial efficiency or the potential or likelihood that an investment made after the design and construction of an additional capacity will result in a higher return on investment or that the return on investment will be more with the lesser amount of capital.\colorbox{yellow!30}{Yellow}
    \item[4] \textit{Asset Turnover Ratios}: Evaluate of  the company's resource usage efficiency. \colorbox{cyan!20}{Blue}
    \item[5] \textit{Market value Ratios}: estimate of  a company's financial standing in the market \colorbox{red!20}{Red}

\end{itemize}

Finally, an extensive review of quantitative fundamental analysis used in research is elucidated in Table(\ref{table4}).

\subsection{Qualitative  Data-Based Studies}
\vspace{6pt}
Qualitative data analysis in the context of stocks involves examining non-numerical information about companies or the market to make investment decisions. This typically involves analysis of text data such as news articles, tweets, financial forum discussions, etc. Multiple studies have shown that The stock price can be significantly impacted by financial news \cite{cutler1988moves}\cite{xie2013semantic}.
Some text mining methods have been summarized in \cite{kumar2016survey}.
Investigations have been performed, utilizing sentiment analysis to assess the influence of sentiments on stock market volatility. Also, it was shown that the Volatility Index(VIX) is highly correlated with market sentiment\cite{bandopadhyaya2008measures}. Preceding research analyzed the role of Author tone from financial articles and evaluated its sentiment\cite{schumaker2012evaluating}. A combined effect of web text news articles and social media sentiment was taken into account in \cite{li2014media}.

Social media has been used extensively by numerous research works through time.\cite{mittal2012stock}\cite{zhang2011predicting}. Tools such as Tweety, OpinionFinder, etc can be used for the extraction and analysis of Twitter data. It's proven already that Twitter's mood impacts stock market prediction\cite{bollen2011twitter}. In \cite{oliveira2017impact}, sentiment and survey indices( USMC, AAII, and Sentix) were extracted from large Twitter data for the S\&P500 index. The Kalman filter was utilized to integrate survey sources and blogs. Finally, for metrics, Diebold-Mariano test was performed to check for a significant difference. 

Initially, some studies used the bag of words approach for embedding news text data for prediction, but Xiao Ding in \cite{ding2014using} showed that, Unlike the bag of words approach, structured events are more helpful depictions for the purpose of making stock market predictions.
He also concluded that in the study of forecasting, the quality of the information is more significant than the amount of data available.
Moreover, studies have investigated the fusion of event and sentiment mood data for prediction. In \cite{li2016tensor} sentiments and events are integrated and exhibit combined influence on the stock market, making use of the tensor framework. 

A recent work by Zhang .et al \cite{ZHANG2018236}  effective integration of events and mood data was implemented addressing the sparse data problem. 
Studies in Behavioural Finance that correlate with investor decision-making processes have discussed the implications of overconfidence \cite{bertella2014confidence}. According to earlier studies on overconfidence and behavioral psychology \cite{skala2008overconfidence} it was stated that investors choose private knowledge over public \cite{ackert2009behavioral}. Thus, because of investors' access to social media information, overconfidence may be reduced. 
Added to this, a comprehensive review of Qualitative Data-Based Studies is shown in Table(\ref{table5}).

%% file: Sections/Dataset_Description.tex
\section{Dataset} \label{dataset}
\vspace{5pt}
In this section, we propose a Huge stock market forecasting dataset consisting of features in three categories: Technical Indicators, qualitative fundamental indicators, and Quantitative fundamental indicators. We consider eight different stocks shown in Table(\ref{table6}) from DJIA (Dow Jones Industrial Average) and provide data for them. Data is also provided for the de facto DJIA index. It is to be noted that Dow Jones is a price-weighted Index. 

The reason for the selection of the specified stocks is to cover a wider range of the market based on the following factors: Industry and Index weighting. A wide range of Industries have been covered in the list. The purposeful inclusion of United Health (Healthcare) is to check for the influence of COVID-19 on healthcare stocks. Further, Boeing (Defense and Aerospace) was selected to compensate for the stock market volatility due to the ongoing Russia-Ukraine War. 

Next, There is considerable variation in the Index weighting for the companies. The highest being United Health with 10\% and the least being \%0.57 by Intel. The purpose of this was to cover a wider range of the market based on the market cap and price. The data provided for each company can be used for training ticker-specific predictions. All data collection was performed from January 2018 to December 2022 for 5 Years. The reason for the collection of data from absolute open source data sources and self-designed extraction tools is to incorporate ease of data access and inexpensive finances for the research and future work. This should be of benefit to the peer research community.




\begin{table*}[h]
\centering
\captionsetup{justification=centering,margin=2cm}
\scalebox{0.82}{
\begin{tabular}{cccc}

  \toprule
\textbf{Stock} & \textbf{Ticker} & \textbf{Industry} & \textbf{Index Weighting(\%)} \\
    \midrule

    Chevron (NYSE) & CVX & Petroleum & 3.50 \\
    Goldman Sachs (NYSE) & GS & Financial Services & 7.36 \\
    Microsoft (NASDAQ) & MSFT & Information Technology & 4.88 \\
    Nike (NYSE) & NKE & Clothing Industry & 2.13 \\
    Walgreens Boots Alliance (NASDAQ) & WBA & Retailing & 0.79 \\
    Boeing (NYSE) & BA & Defense & 3.36 \\
    United Health (NYSE) & UNH & Health Care & 10.29 \\
    Intel (NASDAQ) & INTC & Semiconductor Industry & 0.57 \\

\hline
\end{tabular}}
\caption{Selected stocks from DJIA index}
\label{table6}
\end{table*}

\subsection{Technical Indicator features}
\vspace{5pt}
Experts have also noticed that some groups of indicators are better suited and much more efficient at collecting particular economic/market patterns. For example, the relative difference in percentage (RDP) and moving averages (MA, EMA) are preferred for current high-trend market segments (bullish or bearish), whereas Commodity Channel Index (CCI), BAIS, price oscillator (OSCP) oscillators, true strength index (TSI), \%D (moving average of \%K), and stochastic \%K are better at retrieving market data regarding price. To provide the algorithms with a comprehensive range of indications from which to choose the appropriate one, we employed 45 technical indicators that had been substantially used in earlier research\ref{table1} with multiple input variations.
 

\subsection{Quantitative Fundamental Analysis Features}
\vspace{5pt}
We added Financial Variables and other Macroeconomic Variables from \ref{table3} using various sources for each stock under consideration. The purpose of Quantitative fundamental analysis as mentioned earlier is to assess the financial health of these stocks through Financial firm-specific data. Data comprises company  Income statements, Balance Sheets, Cash flows, Commodities, Macro variables, multiple financial ratios, International currency exchange rates, and other indexes.  In total, we extracted around 220 features, including technical and fundamental qualitative analyses for each stock. Since Dow Jones data does not include company-specific fundamental data, it consists of around 120 features without the inclusion of Qualitative Analysis.

\subsection{Qualitative  Fundamental Analysis Features} \label{qual_datamine}
\vspace{6pt}
\subsubsection{Data Collection Module}
\vspace{4pt}
In the world of big data, there is a large amount of data available on the web. Extraction can be performed through various techniques. There are pre-designed API's for this task. However, APIs can only be used to extract data through queries to a specific data provider/website. So, in our work, we design a personalized system for data extraction from multiple sources. We develop a Web Scraper for data collection from \href{https://archive.org/}{Internet Archive}. This provided us with quality data per day from January 2018 to December 2022 for each year. The data was stock-specific to all eight stocks and lastly "Dow Jones" specific. The data collected was of two types: TV news captions data and Radio Transcripts data. Web scraping was hedged against IP blocking utilizing IP rerouting and random interval multi-threading using Redis-DB. The respective search engine is made open source at \href{https://github.com/batking24/Huge-Stock-Dataset}{Github}.

\begin{figure}[h!]
\includegraphics[width=\linewidth]{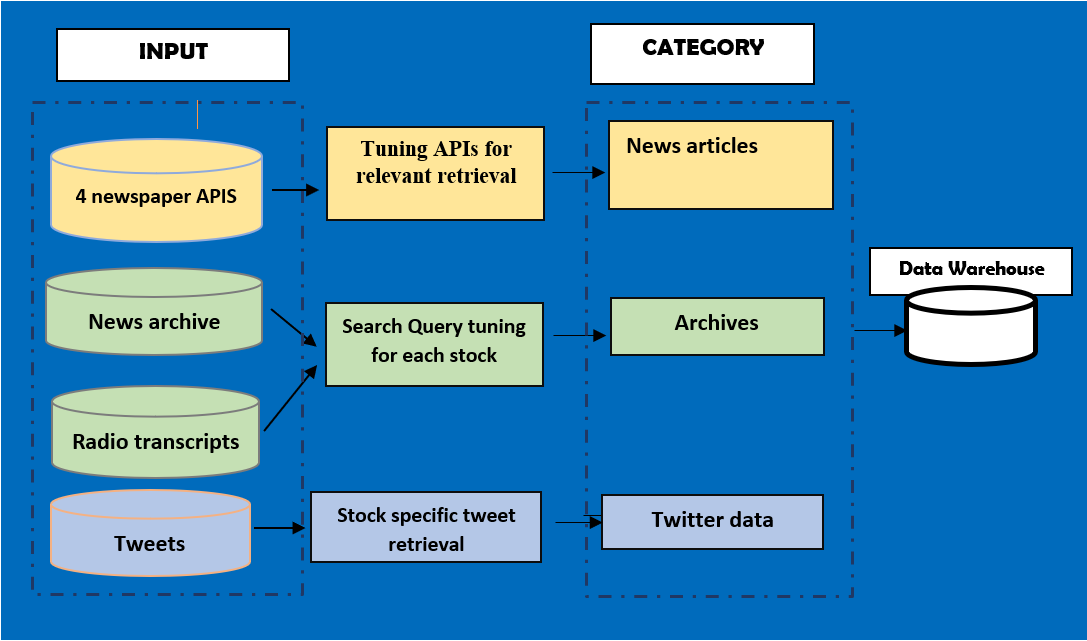}
\caption{Data Retrieval Architecture}
\label{fig:1}
\end{figure}

\begin{figure}[h!]
\centering
\scalebox{0.75}{
\includegraphics[width=\linewidth]{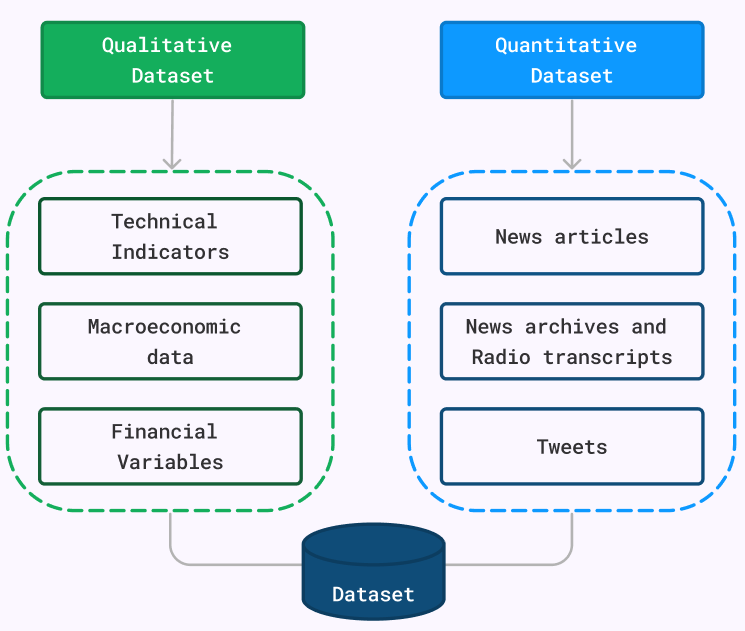}}
\caption{Qualitative and Quantitative data}
\label{fig:2}
\end{figure}

Next, we used the New York Times financial newspaper API, which provides official news articles. The reason for choosing, this API is essentially because NYT provides free access and the API provides date-annotated news articles. Data can also be extracted from various other web news providers like the Financial Times, Thomas Reuters, Refinitiv, and Washington Post.  This provided us with pre-structured data for daily news in the same time frame. However here we encountered some of the most challenging problems with the research which was to find APIs with the capability to query news articles in a specific time frame for a \textit{specific stock} in the \textit{category of Business or Fianance}. Information that does lead to intelligence is useless. This would also confuse machine learning models trained on our data after sentiment analysis was performed on news articles unrelated to a specific stock. The problem with the New York Times was that even though we could extract about 2k large articles, the data wasn't recognized for a specific stock. To work around this we try keyword matching shown in \ref{sec:keywordmatch} but with failure. Hence, we made use of 3 other APIs: Guardian API, Current API, and Alpha Vantage news API. We manually tweaked the parameters of the API for the best retrieval of most related responses in terms of stock entity similarity. Even though these APIs didn't contribute to consistent data(was sparse) through the five years they do provide us with valuable news articles in a more scattered fashion. Next, IP blockage was a challenge for web scraping for radio transcripts and news archive data.  To work around this, we developed robust, privacy-aware retrieval methods that leverage  IP rotation via proxy servers, random time interval scheduling, and user-agent switching using Redis cache to replicate human behavior. 

Finally, we utilize Twitter data. A self-designed Twitter API based on Selenium and SQLite3 was utilized for date and keyword-specific data extraction. The purpose of the code preparation was to utilize as many attributes as possible and get intelligent task-specific data instead of inexact data. This solved the problem of time-specific and stock-specific information retrieval. The code also works with multiple threads and hedges against IP blocking.  In \cite{karlemstrand2021using} it was concluded that Twitter attributes such as likes, followers, retweet number, and account verification add to better predictability of models. Hence, we also capture likes, comment numbers, and retweet numbers along with the post date and Twitter body. We used each stock name and Dow Jones as a query. We capture \textit{0.83 Million} Tweets in total. Twitter extraction engine is available at \href{https://github.com/batking24/Huge-Stock-Dataset}{Github}.

In total, we collected more than \textit{1.4 Million data entries} in total inclusive of all tweets, archive data, and news articles after redundant information was removed and Deduplication was performed.  The final Data Pipeline can be seen from \ref{fig:1}

\subsubsection{Keyword matching} \label{sec:keywordmatch} 
\vspace{6pt}
\vspace{-1 mm}
The structure of the response from NYT API is as follows: Snippet, Web URL, Lead Paragraph, Headline, Keywords, and variables such as type of material and publish date. We extracted 45773 articles in 5 years. Our focus will be on keywords, NYT already provides us with extracted keywords from each news article, however, there is no generic standard form for these words. Hence, to co-relate the news articles with each ticker from the ones selected in \ref{table6} we have to match them. For this, we utilize cosine similarity and edit distance, an approach taken in \cite{zhong2021s}.  After capturing different keywords for each ticker, we then extract stock-specific data per ticker. To capture articles related to Dow Jones directly, we took keywords related to "Dow Jones".  The results were very sparse with only 21 articles from 45773 being extracted. 
The result for each stock index were sparse too as expected\cite{zhong2021s}. We could only get 212 articles that were related to any of the eight stocks from 5 years.
Thus, keyword matching proved to be insufficient. 

\subsubsection{Preprocessing}
\vspace{5pt}
For quantitative data, we don't apply any preprocessing because we want to make the dataset as flexible as possible so that research can apply any type and number of techniques to the original untreated dataset. 

\begin{figure}[h!]

\scalebox{0.55}{

\begin{tikzpicture}[>=latex']
    \tikzset{block/.style= {draw, rectangle, align=center,minimum width=2cm,minimum height=1cm},
    rblock/.style={draw, shape=rectangle,rounded corners=1.5em,align=center,minimum width=2cm,minimum height=1cm},
    input/.style={ 
    draw,
    trapezium,
    trapezium left angle=60,
    trapezium right angle=120,
    minimum width=2cm,
    align=center,
    minimum height=1cm
},
    }
    \node [rblock]  (start) {Whole text Corpus};
    \node [block, right =1cm of start] (acquire) {Remove  punctuation,\\brackets};
    \node [block, right =1cm of acquire] (rgb2gray) {Remove Unicode,\\URLs};
    \node [block, right =1cm of rgb2gray] (otsu) {Remove hashtag words,\\stop words};
    \node [block, below right =2cm and -1.5cm of start, fill=blue!20] (gchannel) {Replace Slang \\ Contractions};
    \node [block, right =1cm of gchannel, fill=blue!20] (closing) {Spelling \\ correction};
    \node [block, right =1cm of closing] (NN) {Lemmetization};
    
    \node [block, right =1cm of NN, fill=green!20] (limit) {Stemming};
    \node [ellipse, draw, right =1cm of limit] (end) {END};
    \node [coordinate, below right =1cm and 1cm of otsu] (right) {};  
    \node [coordinate,above left =1cm and 1cm of gchannel] (left) {};  

    \path[draw,->] (start) edge (acquire)
                (acquire) edge (rgb2gray)
                (rgb2gray) edge (otsu)
                (otsu.east) -| (right) -- (left) |- (gchannel)
                (gchannel) edge (closing)
                (closing) edge (NN)
                (NN) edge (limit)
                (limit) edge (end) {};
\end{tikzpicture}
}

\caption{Preprocessing Pipeline}
\label{fig2}
\end{figure}
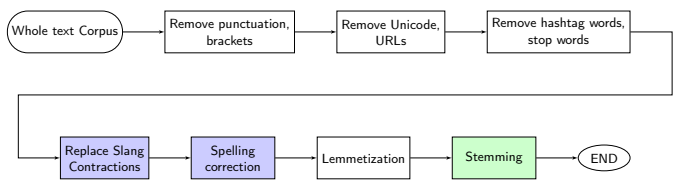

However, for Qualitative data, it is vital that optimal preprocessing of text is performed. This is even more crucial with Twitter data where informal lingo is used by the users. Misspelled words, slang, New words, URLs, etc are common \cite{10.1007/978-3-319-09339-0_62}. Hence, additional preprocessing has to be applied for Twitter data over news and archive data. Further, stemming is to be performed for all news and archive data when used with Loughran-McDonald Master Dictionary Sentiment Analysis. Loughran-McDonald method is elucidated below. This is because all different forms of words might not be present in the dictionary, so it's optimal for all the words to be converted to root form. This way when direct String comparisons are done the words are matched with the dictionary appropriately. For example, the word "boom" is present in the dictionary, but "booming" is not, and unless stemming is used this conversion does not happen. The pipeline used for preprocessing is shown in Figure\ref{fig2}. It is to be noted that we don't remove "@organization" words such as "@Apple" but only remove the "@" token from the corpus. Corpus is also not converted to lowercase, this is to prevent the loss of user intonation and speech intensity information in text. 

Blue and green refer to the techniques used exclusively for Twitter corpus and Loughran-McDonald  Sentiment respectively. Replacing contractions is used to convert words such as won't and didn't to will not and did not. This would be beneficial essentially for Twitter data. However, contraction removal is not performed for the Loughran-McDonald  Sentiment because we compare the preceding single-word string for the negation of sentiments. As mentioned, the usage of informal slang is common in tweets, to replace slang we use a Slang Map dictionary that gives meaning to slang words. \href{http://norvig.com/spell-correct.html}{Spelling correction} is also performed based on edit distance and probability theory.  

\subsubsection{Sentiment analysis tools }
\vspace{5pt}
The purpose of using already trained packages and tools would be for the quick execution and implementation of sentiment calculation. Even though the performance of such techniques might not be cutting-edge, it does provide a baseline output, which could lead to significant features in the dataset. Typically, these consist of two types: Rule-Based and Embedding-based techniques. 

\textit{NLTK VADER sentiment analysis} is an NLP algorithm that works with a combination of lexicon, grammatical rules, and syntactical conventions. It measures both the polarity of the text and its intensity. The lexicon dictionary used is not only comprised of word sentiments but also phrases and emoticons. This would make such a technique ideal for social media text and more informal text sentiment analysis. Grammatical rules are established with heuristic rules that handle adverbs, conjunctions, punctuation capitalization, etc. An output attribute called compound presents the sum of valence scores for each word in the lexicon and shows the sentiment degree rather than the value. Compound values for positive sentiment are >=0.05, neural is [-0.05, 0.05] and negative is <=0.05. 

\textit{TextBlob} from NLTK also utilizes a similar approach to Vader but TextBlob assesses the final sentiment of a sentence by taking a pooling operation such as the average of polarities. This, however, might have a negative effect when there are an equal number of positive and negative words in the text sentence. TextBlob also provides subjectivity, which quantifies the amount of factual information and subjective opinion present in the text. If the texts have greater subjectivity, personal opinion has taken the place of objective knowledge. One more key difference is that TextBlob is also suitable for non-social media text articles. Flair is another tool that uses word embedding classification for the generation of sentiment. Flair was trained on the IMBD dataset. 

We employ all three of these tools despite the common notion that VADER generally exhibits better performance. This is due to the distinctive capacities of each in capturing various dimensions of sentiment.

\subsubsection{Sentiment Analysis using Sentiment Dictionaries}
\vspace{5pt}

We apply  \href{https://sraf.nd.edu/loughranmcdonald-master-dictionary/}{\textit{Loughran-McDonald}} context-specific lexicon for sentiment analysis of the preprocessed text. It was made from yearly reports made public by businesses and contains 184 constricting words, 19 strong modals, 27 weak modals, 2,355 negative, 297 unsure, 904 litigious, and 354 positive terms. The analysis is done for the whole corpus. However, this step works better with formal data such as news articles as compared to informal tweets. Based on the context of services in the Finance Industry, each term in this lexicon is given a straightforward value. Hence, it was developed for Finance-related text processing. It is to be noted that a check for negation is also performed manually in this method, for up to three preceding words when a negative word is encountered. Negative sentiment is allocated when this happens. For example, "The stock market is not doing well" is given a negative polarity and not a positive one.   

Polarity and Sentiment are allocated using the following formulas:

 

\begin{equation} 
\label{eq1}
\begin{split}
	Polarity & = \frac{pos -  neg}{pos +  neg} 
\end{split}
\end{equation}

\begin{equation} 
\label{eq1}
\begin{split}
	 Subjectivity & = \frac{pos -  neg}{count} 
\end{split}
\end{equation}

\textit{pos} and \textit{neg} is the number of positive and negative words found in the text and \textit{count} is the total number of words in each sentence. However, the Loughran-McDonald dictionary is sparse, and often sentences are not allocated any polarity.

Next, \textit{Harvard IV} dictionary is applied in a similar approach. Harvard IV is a Dictionary with each word tagged with positive and negative attributes, in accordance with the psychological Harvard-IV dictionary as used in the General Inquirer software. Harvard IV is less sparse. These methods are extremely fast and require minimal computational power. Though there is no fine tuning, they are task-specific for financial news.  

\subsubsection{Emotion Analysis with \textit{etf-idf }vectors}
\vspace{5pt}
Emotion Analysis is an effective way to capture sentiment from the user and investor-given information, especially tweets. \textit{Tf-idf }is a well-known method to capture the importance of specific words in a document. A modification for this was performed to capture different emotions from documents. A similar approach was taken by \cite{chun2021using} in 2020. In our approach, we consider emotion \textit{tf-idf} or \textit{etf-idf}. In simple terms, for the term "happy", \textit{tf} refers to the frequency of "happy" in a specific document, \textit{idf} is the inverse computation of a number of documents the word "happy" occurs in. For \textit{etf-idf} instead of making a computation for each word we encounter, we have multiple emotions that act as features, every word is tagged with a set of emotions. When that word is encountered based on its emotions we take its \textit{etf} value for each respective emotion. Hence, \textit{etf(for anger)} is the number of words in the document which have anger emotion. Next, \textit{idf( for anger) }is the number of documents that have anger emotion words. 

The tagging of emotions to each word is done using two different lexicons. \\
The first one is a conjunction of \href{https://saifmohammad.com/WebPages/AccessResource.htm}{\textit{NRC Lexicon}} with gold standard lexicon from \cite{buechel2018emotion}. On thorough inspection, it was found that these two lexicons have emotion tags for the most accepted core emotions. NRC lexicon comprised of  Anger, Anticipation, Disgust, Fear, Joy, Sadness, Surprise, and Trust. However, the preceding lexicon only had Joy, Anger, Sadness, Fear, and Disgust. The purpose of considering two lexicons is to maximize the number of unique words captured. NRC Lexicon comprised 14182 and the second lexicon  13915 respectively. Upon analysis, it was found that 4915 words were different from the preceding lexicon, which were not present in the NRC lexicon. Hence, on nulling non-existing emotion features a concatenation was performed to get a total of 19097 words and eight different emotions in our final NRC lexicon. 







NRC VAD \cite{mohammad2018obtaining} lexicon was also considered for emotion analysis, however, the correlation between VAD values and core emotion values was difficult to establish\cite{bualan2019emotion}. Valence values can be used for direct sentiment correlation except emotion analysis requires specific value clustering as shown in \cite{bakker2014pleasure}. Added to this, the NLTK library used in Section 3.3.4 already makes use of the NLTK VAD corpus. Hence NRC VAD was not utilized.  

Secondly, SenticNet 7\cite{cambria2022senticnet} led to a sentiment analysis lexicon that made making use of unsupervised and reproducible sub-symbolic techniques such as auto-regressive language models and kernel methods. The lexicon was huge consisting of 391560 unique terms. Each term was tagged with 24 emotion features which provides for an extensive emotion analysis tool. The emotions were ecstasy, grief, enthusiasm, terror, serenity, melancholy, dislike, rage, contentment, sadness, joy, responsiveness, bliss, anxiety, annoyance, pleasantness, delight, loathing, anger, disgust, eagerness, acceptance, calmness, fear.

\textit{Preprocessing}: Before the computation of etf-idf values, POS tagging is performed on each document. It is to be noted that NLTK POS tagging is context dependent hence it was executed on minimal prepossessing before stop word word removal in Preprocessing Pipeline\ref{fig2}. After POS tagging, only adjectives, Adverbs and interjections were filtered out in each document. Nouns and verbs were first tested for emotion analysis but resulted in poor etf-idf values; hence were removed. POS tag filtering helps extract minimal essential context from each document, which is valuable for emotion analysis.  \\
Next, etf-idf vectors were computed separately for NRC and SenticNet emotions separately. This means that the corpus was run twice with two different lexicons for the computation of etf-idf. The formula used for etf (emotion term frequency) for each emotion is as follows: 

\begin{equation} 
\label{eq3}
\begin{split}
	etf(doc,doc) & = sum(f(term with emo, doc))  
\end{split}
\end{equation}
where \textit{etf} of a given emotion for a document (sentence) is the \textit{sum} of frequency(\textit{f}) of all terms with that emotion in that document respectively. 

The formula for idf(emotion inverse document frequency) for each respective emotion is as follows:

\begin{equation} 
\label{eq4}
\begin{split}
	 idf(emo,doc) & = log\frac{N}{1+df(emo)} 
\end{split}
\end{equation}

Where df(emo) is the number of documents in which the respective emotion words are present. N is the number of documents. Since we calculate emotions per day, this would be all the documents extracted for that specific day.  One is added in the denominator to see that df values lie in the Log domain.   The same formulas for etf-idf are used for NRC and senticnet with each emotion feature set. The final dataset however does not include senticnet features, only eight emotions were included. This is because of the excessive dilution of emotions in sentences due to too many features. 

\subsection{Sentiment Analysis using Pre-Training of Language models}
\vspace{5pt}
Because of the specialized vocabulary employed in the financial arena, general-purpose models are insufficiently effective. Added to this preserving global context, and passing context through the text article is necessary for better sentiment results. Pre-trained language models, which need fewer labeled instances and may be progressively trained on domain-specific corpora, may be able to help with this issue, which we shall test in this section.

We applied transfer learning using various BERT architectures in our work. This was done by using continued pre-training over already pre-trained BERT architectures to transfer the models to perform better for our domain-specific task. Our hypothesis is that further pre-training to our extensive dataset should bring considerable performance improvement for polarity calculations. One feature of the transfer learning techniques is the capacity to continue pre-training of LMs using domain-specific unlabeled corpora. Consequently, the model can pick up on semantic relationships. In the desired domain's written content, which is presumably distinct from a generic corpus in terms of distribution. This strategy has particular promise for a specialized field like finance, where the terminology and language are vastly different from those in a general field.

We answer the following research questions in this section:

\begin{itemize}
    \item[(RQ1)] How does DistilBERT and DistilRoBERTa sentiment performance compare against already existing BERT and FinBERT?

    \item[(RQ2)] Does Pre-training of BERT over huge financial text data assist performance improvement(case of FinancialBERT) against supplementary pre-training? 
    \item[(RQ3)] Does pre-training over our collected domain data help with sentiment classification performance?
    \item[(RQ4)] How does our collected Domain Specific Pre-training help with Spearman correlation with stock returns?  

\end{itemize}

\subsubsection{Dataset}
\vspace{4pt}
We integrate two standard financial sentiment analysis datasets for our task. 
\href{https://huggingface.co/datasets/financial_phrasebank}{\textit{Financial Phrasebank}} comprises 4845 phrases in total in English that were randomly chosen from financial news articles in the LexisNexis database\cite{malo2014good}.
Financial Experts with backgrounds in business and finance annotated these sentences. The annotators were instructed to assign labels based on how they believed that the data in the phrase might impact the stock price of the listed firm. The collection also contains statistics on the degrees of agreement amongst annotators on certain sentences. We only utilize degrees with \textit{66 percent agreement}. This dataset has 4217 labeled rows.

\href{https://sites.google.com/view/fiqa/home}{\textit{Sentiment FiQA}} For the WWW '18 conference's financial opinion mining and question-answering challenge, a dataset called FiQA \cite{maia201818} was produced. Only the first task that is, the Aspect-based financial sentiment analysis dataset is made use of, which consists of 1,174 tweets and news stories related to finance together with the accompanying sentiment score. The labels of the dataset are regressive with continuous values between [-1,1], with 1 having the highest polarity, in contrast to Financial Phrasebank. To make both the datasets uniform for conjunction, we convert these labels into positive, negative, and neutral. On manual inspection and verification by financial experts, we find that for best results positive is [0.15,1], negative is [-0.15,-1], and neutral is the range in between. Each sample additionally specifies which financial institution is being mentioned in the statement. We don't utilize this information. 

Finally, on integration, we get a total of 5328 labeled rows. The training dataset consists of 3835 rows, test 1066, and validation 427 data entries respectively. The distribution of labels can be seen in Table(\ref{table8})
 We designated 20\% of each phrase as a test set, while the remaining 20\% served as the validation set.
 
\textit{Preprocessing}: Language models that are pre-trained such as BERT eliminate the requirement for preprocessing. This is because they are already accustomed to taking raw full-text inputs. Specifically, it employs a multi-head self-attention system to utilize all of the information in a phrase, including punctuation and stop words, from a variety of angles.

As stated in \cite{araci2019finbert}, to overcome Catastrophic forgetting while fine-tuning BERT-Like models, Slanted triangular learning rate, Gradual Unfreezing, and discriminative fine-tuning are employed with FinBERT, BERT, and FinancialBERT. This applies to both pre-trained and non-pre-trained models in our analysis. The rationale for gradual unfreezing and discriminative fine-tuning is that higher-level characteristics require more fine-tuning in comparison to lower-level ones since language modeling information is largely available at lower levels.

\begin{table}[h]
\centering

\scalebox{0.80}{
\begin{tabular}{cccc}

  \toprule
\textbf{Data} & \textbf{Negative} & \textbf{Neutral} & \textbf{Positive} \\
    \midrule

Train & 605 & 1938 & 1292 \\
 Test & 161 & 526 & 379 \\
 Validation & 81 & 209 & 137 \\
All & 961  & 2770 & 2033 \\

\hline
\end{tabular}}
\captionsetup{justification=centering,margin=2cm}
\caption{Dataset Distribution}
\label{table8}
\end{table}

\subsubsection{Evaluation Metrics}
\vspace{4pt}
Three metrics were utilized to evaluate our sentiment classification models: macro F1 score, Area Under Curve(AUC)(One vs One), and Accuracy. It has to be noted that our respective dataset suffers from high-class imbalance problems, and positive and neutral labels are highly dominant. One vs. One AUC computes the average AUC for each pairwise combination of classes that is feasible. Disregarding the disparity in class when there is class imbalance. The area under the curve is extensively used for the performance of multi-class problems and is a recognized measure\cite{pencina2008evaluating}. Macro F1 average computes F1 scores for every class before averaging them. This too is recommended for imbalanced datasets.

\subsubsection{BERT vs Distil Architectures(RQ1)}
\vspace{5pt}
To answer this research question, we compare various set of BERT architectures. Vanilla BERT fine-tuned and FinBERT\cite{araci2019finbert} (which is already pre-trained on TRC2-financial text data) fine-tuned on  the labeled financial dataset respectively,  DistilBERT and DistilRoBERTa fine-tuned respectively. The results obtained by finetuning the aforementioned models are shown in Table(\ref{table9})

DistilBERT\cite{sanh2019distilbert} is a transformer model that is quicker and smaller than BERT. It was pre-trained on the same corpus under self-supervision with the BERT base model acting as a teacher.DistilBERT was not designed to maximize BERT's performance. Instead, Its goal is to maintain as much effectiveness as possible while reducing the vast size of BERT (and having 110M and 340M parameters, respectively).DistilBERT keeps 97\% of its capabilities while shrinking by 40\% and speeding up by 60\%.

\href{https://huggingface.co/distilroberta-base}{DistilRoBERTa} is the distilled version of the RoBERTa base. The training was similar to DistilBERT. The difference is that RoBERTa sharply performs better than BERT, this was due to different hyperparameters while training. So, presumably, DistilRoBERTa shall perform better. The model includes 12 heads, 768 dimensions, six layers, and 82M parameters in total (compared to 125M parameters for RoBERTa-base). DistilRoBERTa typically performs twice as well in relation to Roberta-base asymptotically.

\begin{table}
\begin{center}
\scalebox{0.80}{
\begin{tabular}{ c  c  c  c }

      \toprule		
  Model &Accuracy&F1-score Macro&AUC(OvO) \\ 
      \midrule
  BERT & 0.84 & 0.81 & 0.95 \\
  FinBERT & 0.85 & 0.83 &0.96 \\
  DistilBERT & 0.86 & 0.84 & 0.95 \\ 
  DistilRoBERTa & 0.88 & 0.87 & 0.97 \\ 
  \hline  
 
\end{tabular}}

\end{center}
 \caption{Performance of Language Models at Sentiment Classification}
 \label{table9}
\end{table}

For FinBERT in terms of variation of performance with Batch size, they perform similarly with a batch size of 16 with the best result. Other hyperparameters for best fit include a learning rate of 2e-5(recommended by BERT), epochs of 4, a warm-up proportion of 0.2, and other parameters defaulting.

BERT fine-tuning underperforms FinBERT as stated in \cite{araci2019finbert} even for the dataset under our consideration. However, the difference between Vanilla fine-tuning and our work is we apply guarding techniques against catastrophic forgetting for BERT architecture, too. This means that under the right conditions, pre-training BERT architecture on domain-specific data does improve accuracy. BERT performs similarly with batch size variation with slightly better results for a batch size of 32. All other parameters for best fit are similar to FinBERT. 

DistilRoBERTa and DistilBERT outperform FinBERT and BERT in performance. DistilBERT outperforms in terms of accuracy and F1-Macro score but lags against finBERT in Area Under Curve. The reason why Distil-based architectures tend to perform better in fine-tuning datasets might be because of the reduction in the number of parameters and layers, resulting in better information encoding and capture.  DistilRoBERTa and DistilBERT performed similarly with batch sizes of 16 and 8 respectively, we utilize batch size 16 for lesser training time and faster computation. Other hyperparameters remain similar for both models which include a learning rate of 2e-5, epochs 5, a weight decay of 0.01, and other parameters defaulting. 

From Table(\ref{table9}) we conclude that DistilRoBERTa has the best performance with fine-tuning without further pre-training over the aforementioned models.

\subsubsection{Effect of Extensive pre-training(RQ2)}
\vspace{5pt}
FinancialBERT is a BERT model for domain-specific language representation that has been trained on huge financial corporations. It has a very similar architecture to BERT. It was trained on 1.8M TRC2 financial news articles, 400 thousand Bloomberg News articles, and a number of corporate reports and earnings calls. The difference between the actual FinancialBERT and our execution is we guard it against catastrophic forgetting by using a Slanted triangular learning rate, Gradual Unfreezing, and Discriminative fine-tuning.
FinancialBERT is available at \href{https://huggingface.co/ahmedrachid/FinancialBERT}{FinancialBERT}. In this section, we try to answer if extensive pre-training of BERT architecture in a financial domain positively affects sentiment classification on fine-tuning. We fine-tune the model on our labeled financial dataset. The results in comparison to other aforementioned BERT architectures are shown in Table(\ref{table10}).
FinancialBERT performs best with a batch size of 32 with a gradual performance decline with lower batch sizes. To check for pre-training effects, we use similar hyperparameters to those used for FinBERT fine-tuning. In other domains where NLP is used research in \cite{gururangan2020don} shows that additional pretraining does improve performance.

From Table(\ref{table10}) we conclude that FinancialBERT outperforms vanilla BERT and already pre-trained FinBERT on our labeled financial Dataset. This shows that extensive pre-training in financial domain-specific data over supplementary pre-training improves performance.

\begin{table}

\begin{center}
\scalebox{0.80}{

\begin{tabular}{ c  c  c  c }

  \toprule
  Model &Accuracy&F1-score Macro&AUC(OvO) \\ 
      \midrule
  BERT & 0.84 & 0.81 & 0.95 \\
  FinBERT & 0.85 & 0.83 &0.961 \\
  FinancialBERT & 0.86 & 0.84 & 0.965 \\  
  \hline  
 
\end{tabular}}

\end{center}
 \caption{Performance of BERT models at Sentiment Classification}
 \label{table10}
\end{table}

\subsubsection{Effect of pretraining on Collected Domain Data(RQ3)}
\vspace{5pt}

Our next research question thoroughly checks the performance enhancement of BERT type models by pre-training it over our collected Domain Data, consisting of around the 1.5 million entries collected by us. BERT architectures come with two types of pre-training: Masked Language Modelling(MLM) pre-training and Next Sentence Prediction(NSP). We could either use only MLM pre-training or both MLM and NSP pertaining to utilizing the data at hand. This was followed by the authors of BERT. However, there would be compromises in computational usage and time complexity. 
It is to be noted that NSP along with MLM typically gives more enhanced results as compared to only MLM. To try all variants, we train vanilla BERT with MLM+NSP pretraining, FinBERT, and FinancialBERT with only MLM and distilRoBERTa with its default MLM pre-training. DistilRoBERTa was originally pre-trained using only MLM.  Part of the reason for implementing only MLM over FinBERT and FinancialBERT was due to the failure of integration of these architectures with existing pre-training NSP source codes. The results and performance is  shown in Table(\ref{table11})

From Table(\ref{table11}) we note that there has been a notable improvement in the performance of FinBERT from 0.83 to 0.85. This shines a light on the efficacy of the collected data and its evaluability toward pre-training in the domain of finance. There seems to be no enhancement in performance in FiancialBERT after domain pretraining and F1 remains at 0.84, this could be because of the already extensive pre-training that was performed over financial data. It is worth mentioning that FinBERT outperformed FinancialBERT after pre-training over our Financial domain data when it failed to do so before(RQ2)

Further, there seems to be a notable loss in the performance of BERT pre-training using (MLM+NSP). The reason for this could be because of context irrelevance learned while Next Sentence Pre-training was performed over Domain Data. The decrement in the performance of DistilRoBERT could be because of the same reason for catastrophic forgetting while Domain MLM pre-training was performed.

\begin{table}

\begin{center}
\scalebox{0.80}{
\begin{tabular}{ c  c  c }

  \toprule
  Model &Accuracy&F1-score Macro \\ 
  \midrule
  BERT & 0.82& 0.78 \\
  FinBERT & 0.86 & 0.85\\
  FinancialBERT & 0.86 & 0.84  \\  
  DistilRoBERT & 0.83 & 0.81  \\
  \hline  
 
\end{tabular}}

\end{center}
 \caption{Effect of BERT pretraining over collected Domain Data}
 \label{table11}
\end{table}

\subsubsection{Spearman correlation of stock returns with Sentiment(RQ4)}
\vspace{5pt}
After collection of data which is specific to each stock and date. We created a pipeline for sentiment analysis using FinBERT pre-trained and fine-tuned on our collected corpus. The categories of data available to us are as follows: tweets, news archive snippets, radio transcripts, and full news articles extracted from 4 APIs. Every data entry is annotated with the date of publishing and respective stock. Sentiment analysis is then performed for data categories separately. The reason for this is to show separate correlations and compare data quality. Further, each respective column output could be taken into a weighted average to enhance final prediction results, giving Autonomy to other researchers making use of the data.

\begin{table}
\scalebox{0.80}{
\begin{tabular}{ c  c  c   }
    \toprule
  
  Ticker &Sentiment Category&Spearman Corr \\ 
  \midrule
  DJIA & Twitter-Sentiment & 0.569  \\
  BA & Twitter-Sentiment & 0.424  \\
  MSFT & Twitter-Sentiment & 0.351  \\
  UNH & Twitter-Sentiment & 0.274  \\
  NKE & Twitter-Sentiment & 0.236 \\
  GS & Twitter-Sentiment & 0.298  \\
  DJIA & News-Archive-Sentiment & 0.632  \\
  BA & News-Archive-Sentiment & 0.320  \\
  INTC & News-Archive-Sentiment & 0.233  \\
 DJIA & Radio-transcript-sentiment & 0.463 \\
 CVX & Radio-transcript-sentiment & 0.224 \\
 MSFT & Radio-transcript-sentiment & 0.151 \\
 DJIA & News-Articles(API)-sentiment & 0.448 \\
 INTC & News-Articles(API)-sentiment & 0.103 \\
 WBA  & News-Articles(API)-sentiment & 0.114 \\
 UNH & News-Articles(API)-sentiment & 0.058 \\
 
  \hline  
 
\end{tabular}}
 \caption{Spearman stock return Correlation  with custom FinBERT sentiment for each stock data category }
 \label{stock-corr}  
\end{table}

The pipeline for sentiment analysis is as follows:

\begin{itemize}
  \item Data preparation: extraction of all data entries for a specific stock and date
  \item Duplication of data corpus( especially required for archived information) 
  \item Preprocessing  of text data (different pre-processing for Twitter data) 
  \item Sentiment analysis with custom FinBERT. 
  \item Emotion analysis( only for Twitter data)
\end{itemize}

Finally the pipeline results in five outputs for each date for each stock respectively. The outputs are \textit{Twitter-sentiment, Twitter-emotion,  News-Archive-Sentiment, Radio-transcript-sentiment, and News-Articles(API)-sentiment}. These outputs are based on different data extraction techniques that were mentioned. Twitter emotion results in 9 different emotion values for each day. 
Further, to show the superiority of data quality and sentiment pipeline, we calculated stock return Spearman correlation with sentiment analysis for each respective day.  Stock return is the relative change in yesterday's stock closing price w.r.t today's closing price. 
The non-parametric Spearman rank-order correlation coefficient is used to evaluate the strength and direction of a monotonic connection between two data sets. It ranges from -1 to +1, where -1 and +1 indicate a precise monotonic relationship. A correlation of 0 implies no association between the datasets. Positive correlations signify that as one variable (x) increases, the other variable (y) also increases, while negative correlations indicate that as x increases, y decreases.\cite{myers2004spearman}. We compute Spearman p values for each respective day with stock return. 

As expected, we achieved remarkable results with correlations going beyond 0.6 in some cases. A wide range of notable results are presented in Table(\ref{stock-corr}) for comparison and analysis. The remainder of p-values along with the complete dataset are accessible open source from \href{https://github.com/Bathini-akash/Huge-Stock-Dataset}{DATASET}. Due to sparse data constraints and lack of data for a few days in some categories, sentiment for these respective days is not computed or is computed as absolute 0. Such values are not considered in correlation calculation with stock return. 

From Table(\ref{stock-corr}) it can be inferred that DJIA data performs considerably better than all other stocks. UNH suffers in data quality, this is essentially due to the high presence of words such as "United Health" in other contexts apart from stock and fiance. Such data would only lead to compromised results. Next, a stock that has a better correlation with Twitter data also does so with News Archive, etc. This shows that data availability plays a huge role in data quality and correlation. Radio transcripts on average have an inferior correlation as compared to all other categories. Finally, more prominent stocks such as MSFT, GS, and BA have better correlations due to more data availability and quality. Once again the reason for the selection of stocks with expected scarce availability of data is for the creation of a holistic dataset at the expense of more proficient results(Section(\ref{dataset}))

Further, we perform a similar correlation analysis with emotion analysis for each day for every stock. The correlation p-values for some emotions are given in Table(\ref{stock-emo}). Emotion analysis was performed only for Twitter data. The idea is that individual investor emotions play a vital role in the macro movement of stocks. We capture nine emotions: anger, anticipation, disgust, fear, joy, sadness, surprise, and trust. Positive emotions such as joy and trust are expected to have a positive correlation with stock return and the rest of the negative emotions are expected to have negative emotions. On correlation check, we find out that, anger, disgust, joy, and surprise have a neutral correlation with returns. From Table(\ref{stock-emo}) the following observations can be made: sadness, anger, and fear have expected negative correlations. Trust, anticipation, disgust, joy, and surprise have neutral correlations. A very similar correlation trend is shown for all stocks for mentioned emotions.

\begin{table}
\begin{center}
\scalebox{0.80}{
\begin{tabular}{ c  c  c   }

  \toprule	
  Ticker &Emotion &Spearman Corr \\ 
  \midrule
  DJIA & sadness & -0.266  \\
  MSFT & anger & -0.133  \\
  MSFT & fear & -0.114  \\
  MSFT & sadness & -0.105  \\
  WBA & trust & 0.048  \\
  UNH & fear & -0.105  \\
  BA & anticipation & -0.01 \\
  GS & disgust & -0.090 \\
  NKE & joy & -0.045 \\
  INTC & surprise & -0.009 \\
 
  \hline  
 
\end{tabular}}

\end{center}
 \caption{Spearman stock return Correlation  with emotions for stock specific twitter data}
 \label{stock-emo}
\end{table}

%% file: Sections/Appendix.tex
\section{Appendix}
The following section elucidates the comprehensive literary review performed in the study through various tables.
\onecolumn
{\footnotesize
\begin{longtable}{ |p{5.5cm}|p{10.9cm}}    \toprule

\rowcolor{blue!25} \emph{ \normalsize Technical Indicator} & \emph{ \normalsize Definition}  \\ \hline
\\
 Accumulation/Distribution \newline Oscillator(A/DO)(ACD)     &  A normalised  momentum Oscillator  measure that looks to see if traders are typically purchasing (accumulating) or disposing (distributing) a particular stock in order to determine supply and demand.   \\ \\

Aroon Indicator (Aroon) &  Indicator used to determine the direction and intensity of a movement along with fluctuations in an stock's price. \\ \\

Average True Range (ATR) &  Measures volatility using the average true range over a given period. \\ \\
 
 Acceleration (AC) &  Indicator which measures the acceleration and deceleration of
price. \\ \\ 

Bias or Deviation Rate(BI)  & Indicator which shows how much the stock price deviates from its moving average over a given period of time. \\ \\

Breadth Indicator (BR) & Computes the volume of rising and declining stocks\\ \\

Bollinger Bands (BB)  & A measure of volatility that depicts prices by drawing an upper and lower band utilizing 2 standard deviations along a specified range of days moving average. \\ \\

Chaikin oscillator (CHO) & Measures Adjusted closing (AD) line of MACD  \\ \\

Commodity Channel Index (CCI) & evaluate the difference between the current price level and the mean price level over a specified time period.\\ \\ 

 Disparity (D) &  Calculates the percentage value of relative position the most recent closing price  has  with respect to a chosen moving average. \\ \\

 Exponential Moving Average (EMA)(WMA) & Similar to the Moving average but gives significantly more weight to the most recent data.  \\ \\

 Hightest(Ht) & Highest closing price recorded in the last \textit{t} trading days.  \\ \\ 

Lowest(Lt) & Lowest closing price recorded in the last \textit{t} trading days.  \\ \\ 

 MA Convergence Divergence(MACD) & Indicator used for revealing relative information between two moving averages of prices such as duration, strength and direction of stock's price.  \\ \\

 Maximum Drawdown (MDD) & Maximum loss seen between a stocks high and low before a subsequent high is reached. A measure of downside risk over a certain time frame 
\\ \\ 

Median price (Mt) & Median price of that day ( average of Hight and Low price)  \\ \\

 Momentum (MOME)  & The rate at which a stock is increasing. It serves as a trend line indicator and is regarded as an oscillator.     \\ \\

Moving Average (MA) &  Statistical Indicator that measures the average variation over time in temporal stock data.\\ \\

Moving Average \%K \%D (MAkd) &  The moving average of \%K (\%D) \\ \\

On Balance Volume (OBV) & A momentum-based indicator that forecasts changes in stock price using  flow in volume. When the closing price is greater than the previous day, the volume is added, and when it is lower, the volume is subtracted. EMA is used on OBV.  \\ \\

Price Change (PC) & Metric that compares a stock price at 2 different points in time\\ \\

Price Oscillator(OP) & Moving average price oscillator \\ \\

Percentage Price Change(PPC) & Percentage metric that compares a stock price at 2 different points in time \\ \\

Probability of winning (PP) & Number of winning trades divided by  total number of trades in a trading period.  \\ \\

 Relative Strength Index (RSI) & An indicator of momentum that looks for overbought or oversold situations. It contrasts the mean loss of down times with the up times in a specific period of time. \\ \\

Relative Difference in percentage(RDP) & Relative difference in percentage for closing price  \\ \\
  
 Rate of Change (ROC)    &  The rate of change for any given variable temporally over a predetermined time frame.   \\ \\

Rate of Return & A crucial measure that captures the profit/loss of a trading strategy  based on the initial investment in a certain time frame  \\ \\

 Stochastic (S)    &  A momentum measure that assesses a stock's closing price in relation to the price bracket across which it traded. \%D is equal to the moving average along 3 days of the \%K, and the \%K represents the current market rate of the currency pair.  \\ \\

Signal line (SL) &  Moving average plotted on MACD indicator \\ \\

 Slow Stochastic (SLOW) & Moving average for a specified days over Stochastic(S) (Two variations SLOWk and SLOWd) \\ \\

Standard Deviation (Std) & Statistical gauge of market turbulence that assesses how significantly prices deviate from the mean price \\ \\

True Strength Index (TSI) & Technical momentum oscillator used to identify trends and reversals. \\ \\

True Range (TR) & Indicator which is taken as the greatest of the following: current high less the current low; the absolute value of the current high less the previous close; and the absolute value of the current low less the previous close \\ \\

Ulcer Index (UI) & Assesses downside risk by evaluating the length and severity of market falls. \\ \\

Ultimate Oscillator (UO) & A range-bound indicator with a value that fluctuates between 0 and 100 \\ \\

The Volume rate of change (VROC) & Indicator used to extract stock volume volatility  \\ \\

Volatility Ratio (VR) & Shows the volatility of a stock' s price \\ \\

Weighted Moving Average(WMA) & Moving average with more weight given to most recent data \\ \\

 Williams \%R (W\%R)   &  An indicator of momentum that tracks overbought and oversold conditions. It does this by comparing a stock's close to its high-low range over a certain amount of time, usually 14 days. \\  \\ 

 \hline

\caption{ Techincal Indicators used in Different studies}\label{table1}
\end{longtable}
}

\setlength{\arrayrulewidth}{0.5mm}
\setlength{\tabcolsep}{12pt}
\renewcommand{\arraystretch}{2.5}

{\footnotesize
\begin{longtable}{ |p{4cm}|p{6.5cm}|p{4.5cm}|  }

\hline
\rowcolor{blue!20} \multicolumn{3}{|c|}{\textit{ \large Technical Indicators}} \\
\hline
\rowcolor{Gray1}
 \normalsize Study and Index &  \normalsize Dataset and Comments &  \normalsize Technical Indicators Used \\
\hline

Deep Learning for Stock Market Prediction Using Technical Indicators and Financial News Articles \cite{8489208} \newline \textbf{Stock/Index(s)}: CVX stock price is used & \textbf{Dataset}: Financial news from Reuter's website dating October 20, 2006 to November 21, 2013 \cite{ding2014using}. CVX historical stock data from the same time-frame is taken \newline \textbf{Comments}:Word2Vec model is used to generate embeddings. Best mean test acc is  56.84 \%. \newline \textbf{Models}: LSTMS and CNN's. & EMA8, EMA20, EMA200, MACD9, RSI14, OBV20, BB21, Sk,Sd, MOME, ROC, A/DO, D5, W\%R. \newline \newline  Only one day's  data is used to predict the next day.
\\ \hline

Deep Learning for Stock Market Prediction Using Event Embedding and Technical Indicators \cite{8541310} \newline \textbf{Stock/Index(s)}:S\&P500 and DJIA  & \textbf{Dataset}:1) S\&P500 with Financial news from Reuter's website dating October 20, 2006 to November 21, 2013 2) DJIA with Reddit WorldNews Channel from 8 June 2008 to 1 July 2016 \cite{12345} 3) DJIA with news headlines from Intrinio .\newline \textbf{Comments}:Open IE concept taken from \cite{ding2014using} is used for event extraction \cite{12333}. GloVe is used for word embeddings. Best test acc is  69.86 \%. \newline \textbf{Models}: LSTMS and CNN's. & Sk,Sd, MOME, ROC, A/DO, D5, W\%R.  \newline \newline Normalization is performed. 30 days, 7 days, and 1-day inputs are concatenated to predict the next day.
\\ \hline

Improving Stock Closing Price Prediction Using Recurrent
Neural Network and Technical Indicators\cite{neco_a_01124} \newline \textbf{Stock/Index(s)}:S\&P500, NASDAQ, AAPL(Apple)  & \textbf{Dataset}: Historical stock data collected from Yahoo Finance for the mentioned indexes with a different time from for each index, all before 2017. \newline \textbf{Comments}: PCA was used for feature extraction of technical indicators. Optimization strategies followed: adaptive moment estimation (Adam), and Glorot uniform initialization. Best mean test acc is  69.86 \%. \newline \textbf{Models}: LSTMS and RNN. & ACD, MACD, CHO, H20, L20, Sk, Sd, VPT, MOME, ROC, VROC, OBV, AC, A/DO, W\%R.  \newline \newline Predictions based on
previous 20 days for one day output using sliding window.  
\\ \hline

Stock Price Prediction using Technical Indicators: 
A Predictive Model using Optimal Deep Learning
\cite{agrawal2019stock} \newline \textbf{Stock/Index(s)}:3 banks from National stock exchange of India(NSE), SBI, HDFC and Yes Bank stocks.   & \textbf{Dataset}: Historical stock data collected from Yahoo Finance for the mentioned stocks.Time frame consisted of 2 years i.e. from 16 November 2016 to 15 November 2018 \newline \textbf{Comments}: Pearson co-relation matrix is computed for all indicators used. Correlation tensors are then used as inputs for DNN after feature extraction from STIs.  The best mean acc was 59.25. \newline \textbf{Models}: O-LSTM & MA3, MA10, MA30, EMA,MACD, Sk, Sd, H, RSI, W\%R, Std.  \newline \newline Predictions based on
previous 30  days batch size.   
\\ \hline

Predicting stock market index using a fusion of machine learning techniques \cite{PATEL20152162}, 2015
 \newline \textbf{Stock/Index(s)}: CNX Nifty and S\&P Bombay Stock Exchange (BSE) Sensex.   & \textbf{Dataset}: Historical stock data collected from obtained from two websites ,\href{http://www.nseindia.com/}{1} and \href{http://www.bseindia.com//}{2} for the mentioned stocks. The time frame was from January 2003 to Dec 2012 \newline \textbf{Comments}:A two stage fusion approach is taken, SVRs are used to prepare inputs (from each STI)  which are used for ANN/Randomforest/SVR  training in the second stage.  \newline \textbf{Models}: Support Vector Regression, ANN, Random Forest & MA10,WMA10, MOME, MACD, Sk, Sd, A/DO, RSI, W\%R, CCI .  \newline \newline Predictions based on
1-10 days, 15 days, and 30 days previous data.  
\\ \hline

Using artificial neural network models in stock market index prediction \cite{GURESEN201110389}, 2016
 \newline \textbf{Stock/Index(s)}: NASDAQ Stock Exchange index   & \textbf{Dataset}: Historical stock data collected for NASDAQ in the time frame  08/06/2005–27/05/2013 \newline \textbf{Comments}:The study explored different variations of ANN's but concluded that MLP outperformed others  \newline \textbf{Models}: MLP, DAN2, GARCH-MLP, EGARCH-MLP  & Input Variables are generated using different variants of ANNs. No Technical indicators were used.   \newline \newline Predictions based on 4 days of data.  
\\ \hline

Proximal support vector machine based hybrid prediction models for trend forecasting in financial markets \cite{GURESEN201110389}, 2016
 \newline \textbf{Stock/Index(s)}: 12 Indexes S\&P BSE, DAX, Hang Seng, Jakarta Composite, KLSE Composite, 	EURONEXT 100, CNX NIFTY, Nikkei 225, NYA Composite, Russell 3000, Straits Times, Taiwan Weighted   & \textbf{Dataset}: Historical stock data collected for mentioned indexes from Yahoo Finance for the period January 2008 to December 2013\newline \textbf{Comments}: Four feature selection techniques were used over 55 variants of technical indicators Random Forest (RF), Linear Correlation (LC), Regression Relief (RR), Rank Correlation (RC). Joint prediction error (JPE) was used  for comparative analysis. \newline \textbf{Models}: Proximal support vector machine (PSVM) & 	55 technical indicators were used. MA, EMA, RDP, Sk, Sd, W\%R, BI, MACD, MOME, ROC, OP, MP, CCI, SL, ATR, UO, Ul, TSI with a range of inputs for the indicators.   \newline \newline Supervised PSVM is used and binary training labels are determined based on the closing price. 
\\ \hline

Efficient stock price prediction using a Self Evolving Recurrent Neuro-Fuzzy Inference System optimized through a Modified Differential Harmony Search Technique\cite{DASH201675}, 2016
 \newline \textbf{Stock/Index(s)}: BSE SENSEX of Bombay stock exchange S\&P500   & \textbf{Dataset}: Historical stock data collected for mentioned indexes in the time frame  02 July 2012–06 August 2014 \newline \textbf{Comments}: The study used first-order Takagi Sugeno Kang type fuzzy with two feedback loops. This is used in a Recurrent fashion by feeding the fuzzy rule back to itself. \newline \textbf{Models}: Self Evolving Recurrent Neuro-Fuzzy Inference System (SERNFIS), RCEFLANN, & MA5, W\%R14, RSI14  \newline \newline Predictions based on 3 days of data (closing price and mentioned STI's).  
\\ \hline

 Technical analysis and sentiment embeddings for market trend prediction \cite{PICASSO201960}, 2019
 \newline \textbf{Stock/Index(s)}: NASDAQ100 index and 20 companies in the industry breakdown of the index.  & \textbf{Dataset}: Historical stock data collected  for twenty different stocks in NASDAQ for intra-day trading and news data from Intrinio API in the time frame 03/07/2017 to 14/06/2018.Data is collected for every 15 min. \newline \textbf{Comments}: Embeddings were generated from news data. Closing price is encoded making it a classification problem. Gini Index was used for feature extraction and SMOTE is applied for balancing. SVM and MLP were then applied. Achieved 80 \% annualized return. \newline \textbf{Models}: Multi-Layer Perceptron, SVM  &  MA, EMA, RSI, BB, Sk, Sd, W\%R, TR, ATR     \newline \newline 
Intra-day trading was predicted in 15 minutes time frame.
 
\\ \hline
 A hybrid financial trading support system using multi-category classifiers and random forest \cite{bao2017deep}, 2018
 \newline \textbf{Stock/Index(s)}: NASDAQ, DOW JONES, S\&P 500, NIFTY 50 and NIFTY BANK  & \textbf{Dataset}:  The data sets used to conduct the experiment are collected from \href{https://www.quandl.com}{Quandl} for the period from January 2007 to December 2015 for all mentioned indexes. \newline \textbf{Comments}: hybrid approach integrates weighted multicategory generalized eigenvalue support vector machine (WMGEPSVM) and random forest (RF) algorithms (named RF-WMGEPSVM)  5-fold cross-validation is used for training but test set is consecutive 500 day data. \newline \textbf{Models}: Random Forest, SVM  & Performance measures: ROR, PP, MDD  \newline MA, EMA, RDP, BI, MACD, MOME, ROC, OP, Mt, ATR, TSI, UO, UI CCI, Ht, Lt, SL,  Sk, Sd, W\%R, MAkd   \newline \newline

\\ \hline
A deep learning framework for financial time series using stacked autoencoders and long-short term memory \cite{PICASSO201960}, 2017
 \newline \textbf{Stock/Index(s)}: CSI300( A share China), DOW JONES (New York stock exchange), S\&P 500 (USA), Nikkei225 (Tokyo),  NIFTY 50 (India) Hang Seng Index (Hong Kong)  & 
 
 \textbf{Dataset}:  The data sets used to conduct the experiment are collected from  \href{http://www.wind.com.cn}{WIND} database and \href{http://www.gtarsc.com}{CSMAR} database and is available at \href{https://figshare.com/articles/dataset/Raw_Data/5028110}{Data}, for the period from  Jul. 2008 to Sep. 2016 for all mentioned indexes. \newline \textbf{Comments}: Proposed approach consisted of three parts: wavelet transforms (WT), stacked
autoencoders (SAEs), and LSTM. SAEs were used for feature generation. \newline \textbf{Models}: Auto-Encoders, LSTM  &   MACD, CCI, ATR, BB, EMA20, MA5 and MA10, MOME6 and MOME12, ROC, W\%R, SMI  \newline \newline 
 
\\ \hline

Short-term stock market price trend prediction using a comprehensive deep learning system\cite{shenshort}, 2020
 \newline \textbf{Stock/Index(s)}: 3558 stocks from Chinese Stock Market. & 
 
 \textbf{Dataset}:  Data was collected for the Chinese stock market (3558 stocks)  from  \href{http://www.wind.com.cn}{WIND} open-sourced Tushare API, web-scraping from Sina Finance web pages and SWS Research website for the period of 2 years from 2018 to 2020. \newline \textbf{Comments}: PCA was applied to select optimal technical indicators. Polarization, Min-max scaling and fluctuation percentage were applied on different indicators. \newline \textbf{Models}:LSTM  & MA10, MACD, MOME10, ROC10, RSI5, WNR9, BI20, MACD signal, MACD Hist, PC, PPC, W\%R9,SLOWk, SLOWd, A/DO, Aroon26, BR26, VR26     \newline \newline One to ten trading days are taken as input.

\\ \hline

\caption{ Techincal Indicators used in Different studies}\label{table2}

\end{longtable}}

{\footnotesize
\begin{longtable}{ |p{5.5cm}|p{10.9cm}|}    \toprule
\rowcolor{blue!25} \emph{ \normalsize Macroeconomic and Financial Variables } & \emph{ \normalsize Definition}   \\ \hline

\rowcolor{cyan!20}
Average Collection Turnover (ACT) & Financial indicator used in accounting to assess how effectively  a business collects receivables from its clients. \\

\rowcolor{cyan!20}
Average Inventory Turnover (AIR) & calculates how long it took to sell  merchandise after having acquired it. \\

\rowcolor{green!20} 
Budget Deficit (BD)  & shows how government expenditure and revenue differ. The capital market is primarily impacted by an increase in the budget deficit.\\ 

\rowcolor{gray!15}
Beta Coefficient(B) & Used to quantify an asset or portfolio's volatility relative to the market as a whole. \\

\rowcolor{pink!35}
Cash Ratio (CashR)  & Evaluate of firms liquidity \\
\rowcolor{pink!35}

Current Ratio (CurrR) & Ratio of companies current assets to current liabilities \\

\rowcolor{cyan!20}
Days' sales in inventory (DSI) &  Measure of how long it typically takes a business to sell out its inventory on average.  \\ 

\rowcolor{cyan!20}
Days' sales in receivables (DSR) & gauge of how long it typically takes a company to get paid after a sale. \\

\rowcolor{violet!15}
Debt/equity Ration(D/E) & Ratio which demonstrates the strength of the capital that is accessible vs the money that is being used. A low D/E score indicates that the credit available was not spent. \\ 

\rowcolor{violet!15}
Debt Ratio (DR) & Inverse of Current ratio, total libailities to total assets \\

\rowcolor{red!17}
Earnings per share (EPS) & A measure of a company profitability which is calculated by dividing the company's net income by the total number of outstanding shares   \\

\rowcolor{violet!15}
Equity Ratio & Determined by dividing the firms's total assets by the total shareholders' equity. The outcome shows how much of the assets investors still have a claim on.  \\

\rowcolor{green!20} 
Exchange Rate (ER) &  Key macroeconomic indicator of a country's economic standing globally and is a measure for assessing its level of international competitiveness. In international commerce, a nation's exchange rate may serve as a gauge of its level of competitiveness on the global stage.\\ 

\rowcolor{cyan!20}
Fixed asset turnover (FAT) & A Gauge of how rapidly a business is collecting its credit-card sales. \\

\rowcolor{gray!10}
Fundamental Valuation Efficiency (FVE) & It reflects to how well the investment return was forecasted. \\
\rowcolor{green!20} 
Gross Domestic Product (GDP) &  shows the total amount of finished products and services supplied in a nation during a specific time period. Real GDP is nominal GDP that has been adjusted for inflation. GDP may also be used to gauge the development and success of a country\\ 

\rowcolor{green!20} 
Inflation (I)&   A widespread increase in price level is referred to as inflation. The CPI, or consumer price index, is one of the most used measures of inflation. The CPI is a gauge of how much the average price of a basket of consumer goods and services has changed over time. Sudden inflation disclosures have a detrimental influence on stock returns as stated in \cite{li2010analysis}  whereas predicted inflation has hardly any  effect.\\ 

\rowcolor{green!20}
Interest Rate (IR) & Macroeconomic variable which determines how high the cost of borrowing is, or high the rewards are for saving \\
\rowcolor{cyan!20}
Investment Turnover ratio (ITR)  & Ratio of cost of goods sold to the inventory. Evaluates a company's earnings in relation to its debt and equity  \\ 
\rowcolor{green!20}
International reserves (IntR) & Reserve money that central banks can transfer to one another. \\ 
\rowcolor{violet!15}
Long-term Debt Ratio (LDR) & Measures the proportion of long term debt utilized for financing assests of a firm. \\

\rowcolor{violet!15}
Long-term funds to fixed asset Ratio (LTFFA) & Measures the relationship ratio between long term funds and fixed assets,  long term funds is divided by fixed assets . \\

\rowcolor{green!20}
Market Capitalization( MC) & MC calculates the total number of stocks traded in the market. MC equities can be classified into small, medium and large cap. \\ 

\rowcolor{yellow!30}
Predicted profit Margin (PPM) &  The margin anticipated upon project closing. \\

\rowcolor{red!17}
Price/Book ratio (P/B) & Financial ratio which determines if the stock's share price accurately reflects its worth. Also called Market to book ratio. \\

\rowcolor{red!17}
Price/Earnings ratio (P/E) & Financial ratio which is a very useful evaluation statistic for determining how appealing the present stock price of a company is in relation to its per-share profits. \\

\rowcolor{red!17}
Price/Sales ratio (P/S) & This ratio determines if the stock's share price accurately reflects its worth. \\

\rowcolor{green!20}
Public Debt (PD)  &  Debt that a centralised authority owes\\

\rowcolor{yellow!30}
Profit Margin ratio (PMR)  & A simple ratio of a firms net income(profits) to revenue.  \\

\rowcolor{pink!30}
Quick Ratio (QR)  & ratio determines how well a company can meet its short-term obligations\\

\rowcolor{cyan!20}
Receivables turnover (RT) & a gauge of how rapidly a business is collecting its credit-card sales.\\

\rowcolor{yellow!30}
Return on assets (ROA) & Financial  ratio which represents the income as a percentage of the total assets or resources of the company. \\

\rowcolor{yellow!30}
Return  on equity  (ROE) & Financial ratio which provides a summary of the efficiency with which the shareholder's money were spent and the profit realized from its investment. Low ROE suggests that the resources of the investor were not spent effectively. \\

\rowcolor{green!20} 
Sector Analysis (SA) & The sensitivity of each industry to economic developments varies. Sector-based company classification will spread out a particular risk. It would be best to invest in businesses that have a poor correlation to one another.\\

\rowcolor{cyan!20}
Total asset turnover (TAT)  & Assesses the efficiency with which a business utilises its resources to produce income or sales\\

\rowcolor{gray!15}
Total income growth (TIG)  & Assesses the overall revenue increase of a firm by subtraction of current revenue from the previous.\\

\rowcolor{green!20}
Trade Balance (TB) & Difference between the monetary value of exports and imports \\

\rowcolor{green!20} 
Unemployment rate(UR) &  characterised as a societal phenomena when certain working-age group citizens are unable to find employment that is in line with their skills and credentials.
  The government reduces interest rates when the unemployment rate goes up , which leads to rise in stock market value.  \\
\hline

\caption{Macroeconomic and Financial Variable Definitions}\label{table3}
\end{longtable}}

\setlength{\arrayrulewidth}{0.5mm}
\setlength{\tabcolsep}{12pt}
\renewcommand{\arraystretch}{2.5}
{\footnotesize
\begin{longtable}{ |p{3.8cm}|p{6.5cm}|p{4.6cm}|  }

\hline
\rowcolor{blue!20} \multicolumn{3}{|c|}{\textit{ \large Macroeconomic and Financial Variables}} \\
\hline
\rowcolor{Gray1}
 \normalsize Study and Index &  \normalsize Dataset and Comments &  \normalsize Variables Used \\
\hline

Combining multiple feature selection methods for stock prediction: Union, intersection, and multi-intersection approaches \cite{TSAI2010258}, 2010
 \newline \textbf{Stock/Index(s)}: Taiwan Stock Exchange (TSE)  & \textbf{Dataset}:Data was collected fromTaiwan Economic Journal (TEJ) for Taiwan Stock Exchange (TSE) database in the time frame 2000 to 2007\newline \textbf{Comments}: Sliding window technique was applied with 24:1, training to testing ratio. Seasonal data was used from months 3, 6, 9, 12 only per year. Feature selction was performed using PCA, GA, decision trees CART. Best mean accuracy was 79\%.    \newline \textbf{Models}:ANN used with different FS tech.& \textbf{Macroeconomic}: US GPD, US UR, US IR, US ER, etc \newline\textbf{ Financial Variables}: QR, IT, CurrR, CashR,etc  \newline \newline 85 different fundamental analysis variables were used.
\\ \hline

Fusion of multiple diverse predictors in stock market \cite{BARAK201790}, 2017
 \newline \textbf{Stock/Index(s)}: Tehran Stock Exchange (TSE)  & \textbf{Dataset}: Historical stock data collected  for mentioned indexes in the time frame 2002 to  2012\newline \textbf{Comments}: The study applies a fusion of meta classifiers like AdaBoost, Bagging, Boosting with different variants of Decision Tree. Best mean accuracy was 83.6\% given by Bagging. \newline \textbf{Models}: Decision Tree, LAD Tree, Rep Tree & \textbf{Financial Ratios}: P/E, B, EPS, ER, DR, PPM, TIG, F P/S, EPS, FVE coverage percent, Percent of growth EPS, Predicted profit margin,   \newline \newline 
\\ \hline

Evaluating multiple classifiers for stock price direction prediction\cite{BALLINGS20157046}, 2015
 \newline \textbf{Stock/Index(s)}: 5767 Publicly Listed European stocks  & \textbf{Dataset}:Data was collected fromm Amadeus Database from “bureau van Dijk” in 2009. 2010 was used for testing.\newline \textbf{Comments}: The study compare ensemble methods vs single classification models.Various fundamental Indicators were taken as inputs along with technical indicators. AUC was used as metric.\newline \textbf{Models}:ANN,Random Forest, AdaBoost, SVM, K-nearest Neighbours, LR & \textbf{Financial Variables}: liquidity indicators, Financial Leverage indicators, profitability indicators. Balance sheets information was retrieved. Assets and liabilities were separately categoriesed and taken as inputs.  \newline\textbf{Macroeconomic}: GDP, UR, I, PD, IntR, BD, TB \newline \newline Multiple financial indicators and variables were used as input. 
\\ \hline

A deep learning framework for financial time series using stacked autoencoders and long-short term memory \cite{PICASSO201960}, 2017
 \newline \textbf{Stock/Index(s)}: CSI300( A share China), DOW JONES (New York stock exchange), S\&P 500 (USA), Nikkei225 (Tokyo),  NIFTY 50 (India) Hang Seng Index (Hong Kong)  & 
 
 \textbf{Dataset}:  The data sets used to conduct the experiment are collected from  \href{http://www.wind.com.cn}{WIND} database and \href{http://www.gtarsc.com}{CSMAR} database and is available at \href{https://figshare.com/articles/dataset/Raw_Data/5028110}{Data}, for the period from  Jul. 2008 to Sep. 2016 for all mentioned indexes. \newline \textbf{Comments}: The macroeconomic variable is the last set of inputs to affect money flow in the stock market \newline \textbf{Models}: Auto-Encoders, LSTM  & \textbf{Macroeconomic:} ER( US Dollar Index is selected, enough to capture the impact from the monetary market to the stock market) and IR(selected the interbank offered rate in each market as the proxy, which are SHIBOR,MIBOR, HIBOR, TIBOR and Federal funds rate in US.)
\newline \newline 
\\ \hline

Enhancement of stock market forecasting using an improved fundamental analysis-based approach \cite{chen2017enhancement}, 2017
 \newline \textbf{Stock/Index(s)}: Multiple semiconductor stocks( ASE, TSSSSSMC, MediaTek )  were taken from Taiwan stock exchange & \textbf{Dataset}:Data was collected from database of the Taiwan Economic Journal (TEJ) for multiple stocks for the time period 31st March 2008 to 30th September 2011 \newline \textbf{Comments}: Bottom up approach is used for fundamental analysis. Financial indicators were normalized and weights were calculated using Grey relation analysis for finding correlation between stocks and its financial indicators. Financial news was taken from news portals of Yahoo and Google, which increased the final accuracy by 4\%. POS tagging was used for Chinese parts of speech using CKPI. Information gain was used as FS for text data. \newline \textbf{Models}: AdaBoost, SVM, GA & \textbf{Financial Variables}: DR, LTFFA, CurrR, QR, FAT, ACT, AIR, TAT, ROA, EPS, CashR. 
\\ \hline

Machine Learning for Stock Prediction Based on Fundamental Analysis \cite{9660134}, 2021
 \newline \textbf{Stock/Index(s)}: S\&P100 with 102 stocks in US  & \textbf{Dataset}:Quarterly data was collected from companies' SEC 10\_Q filings, published quarterly. For missing values average of adjacent values were taken.\newline \textbf{Comments}: Long term prediction analysis is performed for 22 years of data. The output variable was relative returns w.r.t. DJIA. This innovate approach is taken because relative returns over absolute return tend to exclude notable factors that have an impact on larger market which leads to getting actual stock performance deducting  entire market performance. RF regressor is used for FS. \newline \textbf{Models}: Feed-forward Neural Network (FNN), RF  , Adaptive Neural Fuzzy Inference System (ANFIS)  & \textbf{Financial Variables}: P/E, ROA, P/B, EPS, CurrR, etc 
 \newline 
 Few Financial ratios were taken, most of the inputs were percentage changes. 
\\ \hline

\caption{ Macro Economic and Financial Variables Indicators used in Different studies}\label{table4}

\end{longtable}}


\setlength{\arrayrulewidth}{0.5mm}
\setlength{\tabcolsep}{12pt}
\renewcommand{\arraystretch}{2.5}
{\footnotesize
\begin{longtable}{ |p{3.7cm}|p{5.2cm}|p{5.9cm}|  }

\hline
\rowcolor{blue!20} \multicolumn{3}{|c|}{\textit{ \large Qualitative Analysis}} \\
\hline
\rowcolor{Gray1}
 \normalsize Study and Index&  \normalsize Data extraction &  \normalsize Methodology\\
\hline

Stock market prediction: A big data approach\cite{7373006}, 2015
 \newline \textbf{Stock/Index(s)}: 2 companies were selected  \newline  & 
 
 \textbf{Qualitative Dataset}:Mozenda Web Crawlers were used for data collection. Twitter data from twitter search APIs. For real time can be streamed on HDFS in Flume and Storm. Domain specific dictionary is used for sentiment analysis of individual words taking a mean for sentence sentiment.  & 
 
 \textbf{Prepossessing}:Lemmatization, stop word removal, URL removal, removal of duplicates (because of retweet)  Results were shown using Rhadoop.\newline
\textbf{ Feature selection } Hive script is run in HDFS using domain specific dictionary.  \newline 
\textbf{Models}: Logistic regression

\\ \hline

Directional Prediction of Stock Prices Using Breaking News on Twitter \cite{7396858}, 2015
 \newline \textbf{Stock/Index(s)}: 30 stocks from Dow jones Index (DJI) \newline \textbf{Results}:Daily prediction accuracy using news data was 69\%. &
 
 \textbf{Qualitative Dataset} 5 year data (from 2010 to 2014)  with 50000 articles from NASDAQ website using \href{http://www.newprosoft.com/}{Web content Extractor}. Url, time, secure, time, date metadata were stored. \href{http://code.google.com/p/boilerpipe/}{Boilerpipe} was used for content extraction from collected data 75000 tweets were collected for 6 months if any of the 30 DJI stocks were mentioned in hourley basis using 
\href{https://dev.twitter.com/streaming/overview}{twitter api} from March 2014 to September 2014{twitter streaming API}    & 
 
 \textbf{Prepossessing}:white space, stop words, number and punctuation removal, made to lowercase and stemming.  N gram  key word features were extracted by creation of document term matrix using \href{
http://cran.r-project.org/web/packages/R.matlab/index.html}{matlab} (1 gram and 2 gram).       \newline
\textit{Sentiment detection} for each statement was perfomed using java \href{http://sentistrength.wlv.ac.uk/}{SentiStrength} lib (uses predefined sentiment corpus), but with \href{http://www3.nd.edu/~mcdonald/Word_Lists.html}{Loughran and McDonald} fin sentiment dictionaries. Before this, \href{https://opennlp.apache.org/}{OpenNLP} was used for sentence detection from documents. Majority polarity was taken as sentiment.\newline
\textbf{ Feature selection }Chi square was used for FS. 10 fold Cross validation was performed \newline 
\textbf{Models}: SVM classifier.  

\\ \hline

Stock Trend Extraction using Rule-based and Syntactic Feature-based Relationships between Named Entities \cite{8920986}, 2019
 \newline \textbf{Stock/Index(s)}: Myanmar Stock Exchange. \newline \textbf{Results}:Average precision, recall and F-score were more than 97\% respectively &
 
 \textbf{Qualitative Dataset}: 333 sentences were taken relating to Myanmar stocks. The paper focused on NER application for extraction of intelligence from unstructured data from web. Two step process whree Rule-based NER was performed then stock trends are estimated by extracting relationships using rules and syntactic feature based extraction.  & 
 
 \textbf{Prepossessing}:Text data was tokenised,stop words were removed, POS tagging and stemming.   \newline
\textbf{ Feature selection } Entities were location, stock name, date, money, quantity and finally relation gives trend. \newline 
\textbf{Models}: Entity annotaions were performed using Apache OpenNLP. 

\\ \hline

Stock Price Prediction Using News Sentiment Analysis
\cite{8848203}, 2019
 \newline \textbf{Stock/Index(s)}: Stocks from S\&P500 companies \newline \textbf{Results}:Best performance was with LSTM with polarity and LSTM with embeddings (1.12 and 1.17 MAPE respectively). &
 
 \textbf{Qualitative Dataset} 5 year data (from February 2013 to March 2017)  with 265463 articles from International daily news websites. The predictions were not performed for the S\&P500 index directly but results were shown for 5 top companies in the index. For \textit{Metric} Mean Absolute Percentage Error (MAPE) was utilized. MAPE is optimal when direction of error in prediction can be ignored. It also works around considerable  deviation bias as  seen in the RMS error. & 
 
 \textbf{Prepossessing}: Log transformation was performed to reduce high difference between stock prices. Differencing was used to transform stock Price series to stationary series. HTMl tags were removed. If company name was in the title, whole article was taken. Everywhere else 5 sentences neighboring the tile were extracted. NLTK was used to extract the sentiment from this data. \newline 
\textbf{Models and method}: ARIMA was used with meta parameter grid search. Facebook prophet was utilized with changepoint prior scale parameter tuning. These two models need the data to be stationary.
With LSTM 3 approach's were taken, first using sentiment data and the previous closing price. Second, using whole text embeddings generated by taking a linear combination of tf-idf weights with word2vec word embeddings then taking average of all words for document embedding. This is given as input to CNN which is in turn given to LSTM. LSTM was trained with both OHLC and sentiment values.

\\ \hline

Using Deep Learning to Develop a Stock Price Prediction Model Based on Individual Investor Emotions \cite{chun2021using}, 2019
 \newline \textbf{Stock/Index(s)}: Korea stock price index 200 (KOSPI 200) \newline  \textbf{Results}:Daily prediction accuracy using IEs was  95 to 96\% based on input time frame. It was concluded that emotion IEs performed better than simple sentiment in all cases.  &
 
 \textbf{Qualitative Dataset} Microblog data was collected from Paxnet.co.kr, a korean stock market community in the time frame from Mar. 9, 2012, to Dec. 8, 2016 \newline  Korean emotion lexicon made by Sohn et al\cite{sohn2012korean}was used to get emotion corpus words. Positive words included surprise, joy, anticipation. Negative emotions included fear, sadness, disgust, enger. 
 &

 \textbf{Methodology}: Tf-idf concepts were utilised with POS tagging to emotion analysis. tf-idf gives the frequency of a term in document plus the frequency of documents in which the term appears. A similar etf-edf is taken which computes frequency of all terms associated with a particular emotion. Once these IEs are generated(PIE for posiive and NIE for negative), considering the rise and fall of stock prices for prices for t-1 and t-2 days, PIE and NIE are selected (conditional extraction of IE). Finally these inputs were used for training DNN. 

\textbf{Models}: DNN   

\\ \hline

Deep Learning for Stock Market Prediction Using Event Embedding and Technical Indicators \cite{8541310}, 2018
 \newline \textbf{Stock/Index(s)}: DJIA (Dow Jones 30) and S\&P500 \newline  \textbf{Results}:Daily prediction accuracy using event embedding plus techincal indicators was 62.02\%.(proposed) &
 
 \textbf{Qualitative Dataset} Financial headlines that contain company names were collected from October 2006 to November 2013 from \cite{ding2014using}. Reddit WorldNews Channel Event \cite{12345}(top 25 headlines each day) were collected and news headlines corresponding to DJIA companies were exacted from Intrinio.    & 

 \textbf{Prepossessing}: Representations and embeddings were extracted from news headline text data using OpenIE architecture\cite{ding2014using} which gives (Actor, Action, Object) tuple for each headline. \href{
https://nlp.stanford.edu/software/openie.html}{Stanford Open} IE was used. Next word embeddings were generated (len=100) using gloves for each tuple. Finally, Event embedding was performed to get a single event vector correlating the actor, action, and object for each headline.  For numerical data x-score normalization was used.       \newline
\textbf{Feature selection}: multi-step adaptive elastic-net (MSAENet) was used
\newline
\textbf{Models}: A self designed Insatance Net was used.  

\\ \hline

Analysis and prediction in sparse and high dimensional text data: The case of Dow Jones stock market \cite{SERT2020123752}, 2020
 \newline \textbf{Stock/Index(s)}: Prediction of  Dow Jones(DJIA) Index \newline \textbf{Results}: Best accuracy was 70.9\%  &
 
 \textbf{Qualitative Dataset} 4 month data  with 12560 articles from New York Times in the time frame January 2017 to December 2017 and 2854333 tweets from four months.
 \newline
  \textbf{NLP module}:A sophisticated NLP module was designed which had 3 components, Named entitiy recognition, Topic Modeling and Sentiment analysis. For \textit{NER}, Apache Open NLP (based on ML), Stanford CoreNLP (based on CRF)  European Commission OpeNER (based on MLP) were used wich were given text as input. 
  \textit{Topic modelling} was perfomed using LDA to generate topics with keywords and score table per input. 
\textit{Sentiment analysis} was done by taking an integration of Apache OpenNLP, CoreNLP and SetniNEL . These modules take in text documents and provide a sentiment. SentiNEL was designed in the paper and was based on SVM.
  
  & 
 
 \textbf{Processing and method}: News was collected from 4 categories Heading, Sub heading, nationl and international. NER module was used to get 41184 Named Entities form the news data. Each news article can have more than one NE. NE's were stored based on news Id. Date,entity and number of times this entity was in the news article. Topic modeling was perfomed to generate a news topic matrix capturing their relation.\newline
 
\textit{Sentiment detection} was done using the designed module. Twitter data was extracted using the keyword 'North Korea" (because of the war with US at the time) \newline
\textbf{ Feature selection }Chi square was used for FS. 10 fold Cross-validation was performed \newline 
\textbf{Models}: SVM classifier.  

\\ \hline

A BERT-based Sentiment Analysis and Key Entity Detection Approach for Online Financial Texts\cite{9437616}, 2021
 \newline \textbf{Stock/Index(s)}: Sentiment prediction is performed no index is predicted \newline \textbf{Results}:Best sentiment prediction accuracy was 96\%. It is to be noted that RoBERT performed better than BERT for sentiment classification. 
 &

 \textbf{Qualitative Dataset} 2 datasets were taken from 2019,\href{
https://www.datafountain.cn/competitions/353}{2019 CCF BDCI}  and \href{https://www.biendata.com/competition/ccks_2019_4}{2019 CCKS }. These are text data based on chinese stock market. 
For Sentiment Analysis RoBERT was trained( fine-tuned) on financial text data and then classification was performed. NER was applied and then sentence matching was used to detect key entities from the extracted onces for each text data input. Finally, MRC was used to extract key entities from tags. This was done by rephrasing tags as MRC questions and then predicting key entities with RoBERT MRC model developed. & 
 
 \textbf{Prepossessing}: Error symbols, URLS and garbage charectors were removed. NER was used to created a NER list dataset. Tags were computed if there is a tag in the input (NRC task). This was used to train the RoBERT model (fine tuning). FInally fine tuned RoBERT was used for sentiment classification of dataset. 

\textbf{Models}: BERT, RoBERT. 
 \newline \textbf{Metrics}:F-score, Accuracy

\\ \hline

A novel multi-source information-fusion predictive framework based on deep neural networks for accuracy enhancement in stock market prediction \cite{nti2021novel}, 2021
 \newline \textbf{Stock/Index(s)}: 2 stocks in Ghana Stock Exchange (GSE) \newline \textbf{Results}:CNN+LSTM achieved 95.78\%, MLP achieved 91.31\%. &

\textbf{Qualitative Dataset}: Data was of 3 types: Twitter (using Tweepy API), News Headlines  (ghanaweb.com, myjoyonline.com and graphic.com.gh using the BeautifulSoup API) and Discussion Forums (sikasem.org)in the time frame January 3, 2017, to January 31, 2020. For each separated polarity (-1 to 1), separated subjectivity (0 to 1)  and sentiment were taken. Stock data from \href{https://gse.com.gh}{GSE}, Macroeconomic indicators from \href{http://www.bog.gov.gh/}{Bank of Ghana} and Google trends index (GTI).& 
 
 \textbf{Prepossessing}:Text data was tokenized, segmented, normalized, and freed from noise. NLTK used for removing symbols, punctuations etc. Fusion of data was done based on Stock ID and date. For tweets, spread of news (counts and favorite tweets) was taken. For all text data sentiment was calculated using NLTK. \newline
\textbf{ Feature selection }CNN with one 64 filters and kernle size 2  \newline 
\textbf{Models}: CNN and LSTM. 2 LSTM layers were used with ADAM (lr=0.001)

\\ \hline

A hybrid model integrating deep learning with investor sentiment analysis for stock price prediction, 2021
 \newline \textbf{Stock/Index(s)}:= Shangai Stock Exchange (SSE) \newline \textbf{Results}:Daily prediction accuracy using news data was 0.0449 MAPE(LSTM+CNN) on average. MAPE was used as \textit{Metric} &
 
 \textbf{Qualitative Dataset} Data was collected from Eastmoney.com from Jan. 01, 2017, to Jul. 31, 2019, using web scraping with Beautiful Soup. For training and testing of Sentiment analysis CNN, ChnSentiCorp ( a Chinese Sentiment corpus was used). Technical indicators and historical data were retrieved and computed from DataYes API.    & 
 
 \textbf{Prepossessing}: Text data was Chinese data. So first text segmentation into words was performed. Stop words were removed. Word embeddings were generated using Word2Vec. A CNN sentiment classification model was trained and developed.      \newline
\textit{Sentiment detection}CNN sentiment detection with 128 filters for each text from word embedding inputs. \newline
\textbf{ Feature selection }GA was used. 10 fold Cross-validation was performed \newline 
\textbf{Models}: MLP, CNN, LSTM, LR, SVM.  Sentiment from CNN model along with numerical data was used as an input to the LSTM model. Other models were compared with CNN sentiment analysis+LSTM.  

\\ \hline

\caption{ Qualitative Sentiment analysis used in Different studies}\label{table5}
\end{longtable}}
\twocolumn